\documentclass[letterpaper]{article} 
\usepackage{aaai25}  
\usepackage{times}  
\usepackage{helvet}  
\usepackage{courier}  
\usepackage[hyphens]{url}  
\usepackage{graphicx} 
\usepackage{natbib}  
\usepackage{caption} 
\usepackage{subfig}
\usepackage{amsmath}
\usepackage{amssymb}
\usepackage{mathtools}
\usepackage{amsthm}
\usepackage{multirow}
\usepackage{fancybox}
\usepackage{changepage}
\usepackage[frozencache,cachedir=.]{minted}
\usepackage{algorithm}
\usepackage{algorithmic}
\usepackage{booktabs}
\usepackage[table, dvipsnames]{xcolor}
\def\mathbi#1{\textbf{\em #1}}

\usepackage{newfloat}
\usepackage{listings}
\DeclareCaptionStyle{ruled}{labelfont=normalfont,labelsep=colon,strut=off} 
\floatstyle{ruled}
\newfloat{listing}{tb}{lst}{}
\floatname{listing}{Listing}
\title{Multi-objective Evolution of Heuristic Using Large Language Model}

\author {
Shunyu Yao\textsuperscript{\rm 1, }\thanks{Equal contribution.}
Fei Liu\textsuperscript{\rm 1, }\footnotemark[1]
Xi Lin\textsuperscript{\rm 1},
Zhichao Lu\textsuperscript{\rm 1},
Zhenkun Wang\textsuperscript{\rm 2, }\thanks{Corresponding authors.},
Qingfu Zhang\textsuperscript{\rm 1, }\footnotemark[2]
}
\affiliations {
\textsuperscript{\rm 1}Department of Computer Science, City University of Hong Kong, Hong Kong, China\\
\textsuperscript{\rm 2}School of System Design and Intelligent Manufacturing, Southern University of Science and Technology, Shen Zhen, China\\

\{shunyuyao8, fliu36\}-c@my.cityu.edu.hk, 
\{xilin4, zhichao.lu\}@cityu.edu.hk,
wangzhenkun90@gmail.com, \\
qingfu.zhang@cityu.edu.hk
}

\begin{document}

\maketitle

\begin{abstract}

Heuristics are commonly used to tackle various search and optimization problems. Design heuristics usually require tedious manual crafting with domain knowledge. 
Recent works have incorporated Large Language Models (LLMs) into automatic heuristic search, leveraging their powerful language and coding capacity.
However, existing research focuses on the optimal performance on the target problem as the sole objective, neglecting other criteria such as efficiency and scalability, which are vital in practice.
To tackle this challenge, we propose to model the heuristic search as a multi-objective optimization problem and consider introducing additional practical criteria beyond optimal performance. 
Due to the complexity of the search space, conventional multi-objective optimization methods struggle to effectively handle LLM-based multi-objective heuristic search.
We propose the first LLM-based multi-objective heuristic search framework, \underline{M}ulti-objective \underline{E}volution \underline{o}f \underline{H}euristic (MEoH), which integrates LLMs in a zero-shot manner to generate a non-dominated set of heuristics to meet multiple design criteria. 
We design a new dominance-dissimilarity mechanism for effective population management and selection, which incorporates both code dissimilarity in the search space and dominance in the objective space.
MEoH is demonstrated in two well-known combinatorial optimization problems: the online Bin Packing Problem (BPP) and the Traveling Salesman Problem (TSP). The results indicate that a variety of elite heuristics are automatically generated in a single run, offering more trade-off options than the existing methods. It successfully achieves competitive or superior performance while improving efficiency up to 10 times. Moreover, we also observe that the multi-objective search introduces novel insights into heuristic design and leads to the discovery of diverse heuristics. 

\end{abstract}

\begin{links}\link{Code}{https://github.com/Optima-CityU/LLM4AD}\end{links}

\section{Introduction}

Heuristics are commonly used in solving optimization and decision-making problems in a variety of fields, including engineering~\citep{bozorg2017meta}, industry~\citep{silver2004overview}, and economics~\citep{vasant2012meta}. Unlike exact methods, heuristics offer practical alternatives for finding sub-optimal solutions within a reasonable time cost~\citep{pearl1984heuristics} and are particularly adept at handling complex problems with diverse attributes and constraints. However, developing effective heuristics typically requires expert knowledge and involves laborious trial-and-error manual crafting, presenting a significant challenge for real-world applications.

To address this challenge, much effort has been devoted to automating the design of heuristics~\citep{pillay2021automated}. These efforts can be broadly classified into three categories: heuristic configuration~\citep{ramos2005logistic,visheratin2016automatic}, heuristic selection~\citep{tang2014population, xu2010hydra}, and heuristic composition~\citep{burke2010classification, drake2020recent, pillay2018hyper}. Despite the successful creation of novel heuristics, the effectiveness of these heuristics still heavily relies on algorithmic components crafted by human experts~\cite{drake2020recent}.

In recent years, Large Language Models (LLMs) have demonstrated remarkable capabilities in algorithm design~\citep{liu2024systematic}. The integration of LLMs with Evolutionary Computation (EC) has enabled the automatic generation and refinement of heuristics along with their corresponding code implementations~\citep{liu2024eoh, romera2024mathematical, ye2024reevo}. The designed heuristics achieved competitive performance with minimized human design and model training. However, all existing LLM-based evolutionary heuristic search methods focus on a single objective regarding the optimized performance of the target problem~\cite{ma2023eureka,nasir2024llmatic,liu2024eoh, romera2024mathematical,zhang2024understanding,yao2024evolve,van2024llamea,li2024auto,zengtngps,mao2024identify,ma2024llm}. Other important heuristic design criteria, such as heuristic complexity~\citep{ausiello2012complexity} and code readability~\citep{buse2009learning}, which could be vital in practice, are often neglected. Although some studies have attempted to optimize multiple objectives by combining them into a single objective function, resulting in a single heuristic, the conflicting nature of diverse objectives often makes it challenging to find a single heuristic that satisfies all simultaneously. The exploration of effective methods for searching a set of non-dominated heuristics in a single run remains unexplored.

In this study, we model the automatic heuristic design as a multi-objective optimization problem~\citep{dreo2009using} and propose the first LLM-based multi-objective heuristic search framework, termed \underline{M}ulti-objective \underline{E}volution \underline{o}f \underline{H}euristic (MEoH), to effectively search for a set of non-dominated heuristics in a single run. The contributions of this paper are as follows:

\begin{itemize}
\item We propose an LLM-based automated heuristic design framework to consider the heuristic design from a multi-objective optimization perspective.

\item We propose a dominance-dissimilarity mechanism to enhance diversity and improve search efficiency by considering both the dominance relationships in the objective space and the dissimilarity of heuristics in the search space.

\item We demonstrate the superiority compared with the counterpart of single-objective LLM-based automated heuristic design on two classical optimization problems: the Traveling Salesman Problem (TSP) and the online Bin Packing Problem (BPP). 

\end{itemize}

\begin{figure*}
\centering
\includegraphics[width=0.76\textwidth]{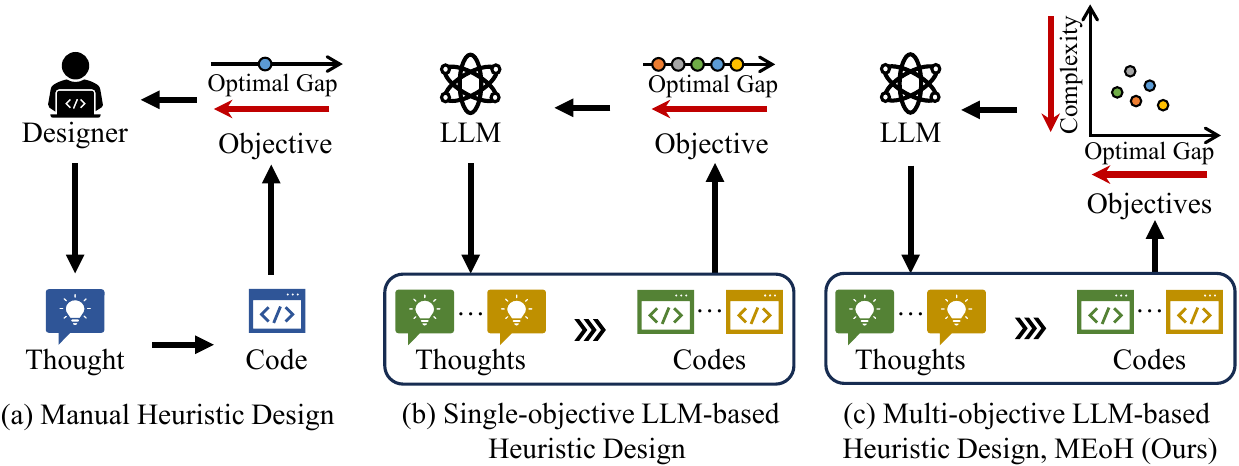}
\caption{Comparison to human design and existing LLM-based heuristic design (a) manual heuristic design by human experts, (b) single-objective LLM-based heuristic design (e.g., FunSearch and EoH), and (c) our proposed multi-objective heuristic design (MEoH).}
\label{fig:MEoH}
\end{figure*}

\section{Related Works}

\subsection{Automated Heuristic Design}

Automated heuristic design methods can be broadly classified into automated heuristic configuration, automated heuristic selection, and automated heuristic composition~\citep{pillay2021automated}. The first category involves using optimization methods and machine learning techniques~\citep{ramos2005logistic,visheratin2016automatic} to automatically adjust the parameters within a given algorithm framework~\citep{agasiev2017program}. The second category focuses on automatically choosing a suitable heuristic for each specific instance from a pool of existing heuristics~\citep{tang2014population, xu2010hydra}. The third category combines various algorithmic elements to create novel heuristics ~\citep{burke2010classification, drake2020recent, pillay2018hyper}. While these methods have shown promise in enhancing the automation of heuristic design and improving performance, they still heavily rely on human-designed algorithmic components.

\subsection{LLM-based Automated Heuristic Design}

Large language models have shown remarkable performance across a variety of tasks and exhibit promising zero-shot capabilities in linguistic processing and code generation. The use of LLMs in automated heuristic design is still in its early stages~\citep{liu2024systematic}. For example, FunSearch~\citep{romera2024mathematical} leverages LLMs to generate and improve code implementations of heuristics based on EC frameworks, achieving state-of-the-art results in mathematical and combinatorial optimization problems. EoH~\citep{liu2024eoh} evolves both idea descriptions and code implementations of heuristics simultaneously, leading to competitive performance in a more efficient manner. This EC+LLM approach has been successfully applied in heuristic and function design across various tasks such as reward function design~\cite{ma2023eureka}, molecular design~\cite{wang2024efficient}, network design~\cite{mao2024identify}, and Bayesian optimization~\cite{yao2024evolve}.  While effective heuristics are developed, they often focus solely on performance for specific target instances, overlooking other crucial objectives like efficiency and complexity.

\subsection{Multi-objective Heuristic Design}

Heuristic design can be modeled as a multi-objective optimization problem. \citet{dreo2009using} consider automated heuristic design as a multi-objective problem to design a set of non-dominated heuristics to balance optimality and efficiency. S-Race~\citep{zhang2013s} employs a racing algorithm to automatically choose machine learning models based on multiple objectives. Furthermore, MO-ParamILS~\citep{blot2016mo} extends the single-objective heuristic configuration framework ParamILS to handle multiple objectives. Multi-objective genetic programming has also been utilized in heuristic search~\citep{schmidt2009distilling, vladislavleva2008order, fan2024decomposition}. However, they still demand existing hand-crafted primitives for defining and generating heuristics.

\section{Preliminaries}
\subsection{Multi-objective Optimization}
\label{sec: MOO}
A Multi-objective Optimization Problem (MOP) can be defined as 
\begin{equation}
\min_{\boldsymbol{x}\in \mathcal{X}} \ \boldsymbol{f}(\boldsymbol{x}) = (f_1(\boldsymbol{x}), f_2(\boldsymbol{x}),\ldots, f_M(\boldsymbol{x})),
\label{eq_moco_problem}
\end{equation}
where $\mathcal{X}$ represents the search space, $\boldsymbol{x}$ is a decision vector, and $\mathbi{f}(\boldsymbol{x})$ is an $M$-objective vector to optimize. A non-trivial MOP cannot be solved by a single decision vector, and we have the following definitions for multi-objective optimization: 

\textbf{Pareto Dominance}: Let $\boldsymbol{x}_a, \boldsymbol{x}_b \in \mathcal{X}$, $\boldsymbol{x}_a$ is said to dominate $\boldsymbol{x}_b$ ($\boldsymbol{x}_a \prec \boldsymbol{x}_b$) if and only if $f_i(\boldsymbol{x}_a) \leq f_i(\boldsymbol{x}_b), \forall i \in \{1, 2, \ldots, M\}$ and $f_j(\boldsymbol{x}_a) < f_j(\boldsymbol{x}_b), \exists j \in \{1, 2, \ldots, M\}$.

\textbf{Pareto Optimality}: A decision vector $\boldsymbol{x}^* \in \mathcal{X}$ is Pareto-optimal if there does not exist $\boldsymbol{x}^\prime \in \mathcal{X}$ dominates $\boldsymbol{x}^*$, i.e., $\nexists \boldsymbol{x}^\prime \in \mathcal{X}$ such that $\boldsymbol{x}^\prime \prec \boldsymbol{x}^*$. 

\textbf{Pareto Set/Front}: The set of all Pareto-optimal decision vectors is called the Pareto Set (PS), and its mapping in the objective space is called the Pareto Front (PF).

In this paper, we investigate multi-objective heuristic design. The decision vector $\boldsymbol{x}$ indicates the heuristic and the $M$-objective vector represents different criteria measuring different aspects of the performance of heuristics (e.g., optimal performance and complexity). 

\subsection{Multi-objective Evolutionary Algorithms}
Multi-objective Evolutionary Algorithms (MOEAs) are among the most commonly used methods to solve MOPs. MOEAs work by maintaining a population of $N$ candidate individuals that evolve iteratively through genetic operators like crossover and mutation. There are three main paradigms for MOEAs: the dominance-based approach~\citep{deb2002fast}, the decomposition-based approach~\citep{zhang2007moea}, and the indicator-based approach~\citep{zitzler2004indicator}.

\section{Methodology}

\subsection{Framework}

Multi-objective Evolution of Heuristic (MEoH) is a fusion of LLMs and multi-objective evolutionary optimization for effective multi-objective heuristic design. As illustrated in Algorithm~\ref{alg:MEoH}, MEoH begins with population initialization, where the population comprises heuristics, and progressively improves the population using MOEA until the termination condition is satisfied, to obtain a set of non-dominated heuristics that represent trade-offs among multiple objectives. Throughout each iteration, MEoH generates offspring using search operators. These operators are implemented through LLMs and predefined prompts to create offspring based on the selected parents from the population. New offspring are added to the population and population management is utilized to update the population to keep its size, with a focus on maintaining diversity and convergence. The dominance-dissimilarity mechanism is utilized in both parent selection and population management. Detailed explanations of each of these components will be provided in the subsequent sections.

MEoH advances existing LLM-based heuristic design by extending the single-objective approach~\citep{romera2024mathematical,liu2024eoh} to the multi-objective scenarios and designing a set of non-dominated heuristics in a single run. Moreover, unlike directly combining MOEA and LLM-based heuristic search, MEoH introduces a unique dominance-dissimilarity measure to navigate the complex and discrete heuristic search space, overcoming challenges faced by conventional MOEAs like NSGA-II~\citep{deb2002fast} and MOEA/D~\citep{zhang2007moea}.

\begin{algorithm}[ht]
\begin{algorithmic}[1]
\STATE \textbf{Input:} Population size $N$; Maximum number of iterations $T$, Parent selection size $d$; Initial population $\boldsymbol{P}_0$; Pre-trained LLM $\mathcal{L}$.
\STATE \textbf{Output:} Approximate Pareto-set $\boldsymbol{P}^*$.\\
\IF{$\boldsymbol{P}_0 = \emptyset$}
\FOR{$i = 1,\ldots,N$}
\STATE $o \leftarrow$ Generation($\mathcal{L}$);
\STATE $\boldsymbol{P}_{0} \leftarrow \boldsymbol{P}_{0} \cup o$
\ENDFOR
\ENDIF
\FOR{$t = 1,\ldots,T$}
\FOR{$i = 1,\ldots,N$}
\STATE $\boldsymbol{P}_{parent} \leftarrow$ ParentSelection($\boldsymbol{P}_{t-1}$, $d$);
\STATE $o \leftarrow$ Search($\mathcal{L}, \boldsymbol{P}_{parent}$);
\STATE $\boldsymbol{P}_{t-1} \leftarrow \boldsymbol{P}_{t-1} \cup o$
\ENDFOR
\STATE $\boldsymbol{P}_{t} \leftarrow$ PopulationManagement($\boldsymbol{P}_{t-1}, N$)
\ENDFOR
\STATE $\boldsymbol{P}^* \leftarrow \boldsymbol{P}_{T}$
\end{algorithmic}
\caption{MEoH}
\label{alg:MEoH}
\end{algorithm}

\subsection{Dominance-dissimilarity Mechanism}

Traditional MOEAs~\cite{deb2002fast,zhang2007moea} and single-objective LLM-based heuristic design methods~\cite{romera2024mathematical,liu2024eoh} lack effective diversity maintenance strategies for multi-objective automated heuristic design. To address this, we propose a novel dominance-dissimilarity mechanism that considers both objective space dominance and heuristic search space dissimilarity.

\paragraph{Dominance Measure in Objective Space:} In the objective space, the Pareto dominance relationship between each pair of heuristics is evaluated, which is widely used in MOEAs~\cite{zitzler1998evolutionary, deb2002fast}. 

\paragraph{Dissimilarity Measure in Search Space:} In the search space, the heuristics are represented through natural language descriptions and implemented in Python code. We evaluate the dissimilarity between code segments. Notably, there are various techniques available for this purpose, and we choose to utilize the widely adopted Abstract Syntax Tree (AST)~\cite{neamtiu2005understanding}. The AST converts the code segment to an abstract syntactic structure~\cite{baxter1998clone}. And the similarity of code $a$ and code $b$ can be calculated based on the tree structures following~\citet{ren2020codebleu}:
\begin{equation}
\text{Sim}_{\text{AST}}(a, b) = \text{Count}_\text{clip}(\text{Tree}_a)/\text{Count}(\text{Tree}_b), 
\end{equation}
where $\text{Count}(\text{Tree}_b)$ is the number of subtrees of $\text{Tree}_b$, and $\text{Count}_\text{clip}(\text{Tree}_a)$ is the number of subtrees of $\text{Tree}_a$ that are matched the $\text{Tree}_b$. The AST similarity value ranges from 0 to 1, with 0 indicating complete dissimilarity between the two code segments and 1 signifying identical code segments. This quantitative approach enables the assessment of structural similarity between code segments, facilitating the comparison and evaluation of heuristics based on their code implementations. 

\paragraph{Dominance-dissimilarity Score:}

\begin{figure*}
\centering
\includegraphics[width=0.70\textwidth]{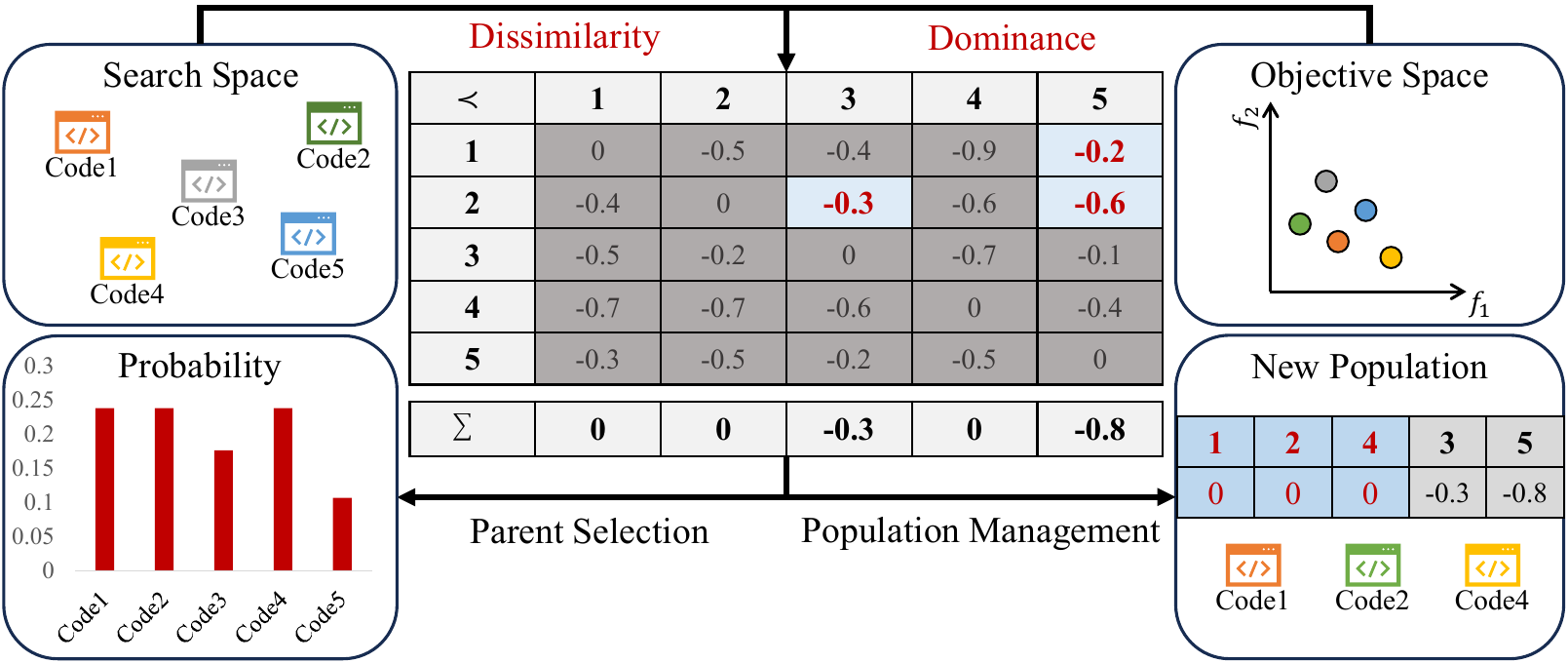}
\caption{An illustration of parent selection and population management with dominance-dissimilarity mechanism. By incorporating code dissimilarity in the search space and dominance relationships in the objective space, the parent selection and population management are enhanced to promote diversity and improve search efficiency. }
\label{fig:dominance-dissimilarity}
\end{figure*}

As illustrated in Figure~\ref{fig:dominance-dissimilarity}, to determine the dominance-dissimilarity of each heuristic in the population, the dissimilarity, i.e., the negative AST similarity, between each pair of heuristics is calculated and stored in a matrix. Concurrently, in the objective space, the dominance relationship between each pair of heuristics is captured and represented as a mask with the same size as the dissimilarity matrix. Specifically, only the dominance relationship is considered, while all other relationships are masked. Subsequently, the masked dissimilarity matrix is aggregated column-wise. The resulting dominance-dissimilarity score vector encapsulates both dominance and diversity aspects to guide parent selection and population management in the subsequent steps. The details can be found in Appendix~\ref{sup:algorithms}. 

\subsection{Heuristic Representation}
Similar to \citet{liu2024eoh}, each heuristic in MEoH is composed of three elements: a description in plain language, a code snippet in a specific format, and a fitness score. 

The description is a brief linguistic explanation generated by LLMs that conveys the main idea. The code snippet is the actual implementation of the heuristic. In the experiments, we opted to use Python functions for implementation. The code snippet must include the 1) function name, 2) input variables, and 3) output variables for clarity. The fitness is evaluated on a set of instances for the specific target problem. Example heuristics can be found in Appendix~\ref{sup:heu_example}.

\subsection{Heuristic Generation}

\subsubsection{Initial Heuristic Generation}
The initial population of MEoH is comprised of heuristics. These heuristics can be generated by leveraging a LLM with a predefined generation prompt or by using human-designed existing heuristics. In order to fully demonstrate the capability of MEoH in designing competitive heuristics, we let LLM generate all the heuristics in both the initiation and evolution processes.

\subsubsection{Offspring Heuristic Generation} 

The parent selection is the first step of generating offspring, in which a set of parent heuristics $\boldsymbol{P}_{parent}$ are selected from the current population. To consider both convergence and diversity in the heuristic search process, the dominance-dissimilarity score is utilized to guide the probability of parent selection. A higher dominance-dissimilarity score indicates a lower likelihood of being dominated or a more diverse code segment, making it preferable. The parents are selected with probability proportional to their dominance-dissimilarity scores. The details can be found in Appendix~\ref{sup:algorithms}. 

The selected parent heuristics serve as samples in the prompt to instruct LLM in generating offspring heuristics. We employ five different search operators with diverse prompt strategies adapted from EoH~\cite{liu2024eoh} to produce offspring heuristics. The details of these prompts can be found in Appendix ~\ref{sup:operator}.

\subsection{Population Management}
As the offspring generated through search operations are incorporated into the population, the size of the population gradually increases. In order to ensure a consistent population size and update the population effectively, a population management strategy is proposed. The dominance-dissimilarity score is utilized for this purpose. Specifically, the heuristics in the population are sorted based on their dominance-dissimilarity score and the worst heuristics are removed to ensure that only the most promising individuals are retained within the population, as detailed in Appendix~\ref{sup:algorithms}. By employing this strategy, the population is continually refined to maintain a high-quality and diverse set of individuals, enhancing the overall efficiency and effectiveness of the evolutionary process.

\section{Experiments}

\subsection{Experimental Settings}

\subsubsection{Problems \& Implementation Details}  

We demonstrate MEoH on two representative combinatorial optimization problems: 

1) \textit{Online Bin Packing Problem: } In online Bin Packing Problem (BPP)~\citep{seiden2002online}, a set of items, each with its own weight, needs to be packed into bins with a predetermined capacity. The objective of the BPP is to minimize the total number of bins required to accommodate all the items. In an online scenario, items are packed as they are received without prior knowledge. The generated heuristics are evaluated on $5$ Weibull instances with $5,000$ items (referred to as 5k), and the capacity of bins is $100$. 

We inherit the settings from~\citet{romera2024mathematical} to design constructive heuristics for aligning the arriving items to the appropriate bins. The designed heuristics involve a function scoring the bins, where the input includes the arriving item size and the remaining capacities of the bins. The item will be assigned to the bin with the highest score. 

2) \textit{Travelling Salesman Problem: } In Traveling Salesman Problem (TSP)~\citep{reinelt2003traveling}, the objective is to find the shortest route that visits all given nodes exactly once and returns to the starting node. In this work, we evaluate the fitness of designed heuristics during evolution on $64$ instances with $100$ nodes. The coordinate of each node is randomly sampled from $[0, 1]$~\citep{kool2018attention}.

The Guided Local Search (GLS) framework is employed~\citep{voudouris2010guided} to iteratively improve the solution quality following~\citep{liu2024eoh}. GLS iteratively performs two steps: 1) local search and 2) perturbation. Until the stop criterion is satisfied, the best solution obtained throughout the iterations is considered the final solution. We aim to design a heuristic to update the distance matrix in the perturbation step. 

The experimental parameter settings are as follows: the number of generations is $20$, and the population size is $20$ and $10$ for online BPP and TSP, respectively. Each crossover operator selects 5 parent heuristics to reproduce the offspring heuristics. The number of iterations and running time in the GLS for TSP is limited to $1,000$ and $60$ seconds, respectively.

\subsubsection{Environments}
To ensure fairness and consistency, all experiments in this study were conducted on a computer equipped with an Intel Core i7-11700 processor and 32GB of memory. GPT3.5-turbo is employed as the per-trained LLM, with each experiment repeated three times to ensure the robustness and reliability of the results. 

\subsubsection{Performance Metric} 

\paragraph{Objectives} 1) \textit{Optimal Gap:} We use the optimal gap to baseline as the first objective (e.g., the gap between the number of bins used in designed heuristics to the lower bound of bin number). 2) \textit{Efficiency:} The running time of heuristics is used as the second objective to reflect the efficiency of heuristics. 

\paragraph{Metric}
1) \textit{Hypervolume: } The Hypervolume(HV) is a commonly used metric in multi-objective optimization. It provides a comprehensive assessment of convergence and diversity of the approximate Pareto front without the ground truth Pareto front~\citep{audet2021performance}. A larger HV value indicates a better performance. 2) \textit{IGD: } The Inverted Generational Distance(IGD) measures the quality of the generated approximate Pareto front in relation to the reference set. Here the reference set is the nondominated set derived from the union of all generated heuristics. A lower IGD value is preferred, which indicates better convergence and diversity, implying that the generated population is closer to the reference set. The detailed formulation of the two metrics can be found in Appendix~\ref{sup:metric_definition}.

\subsubsection{Baseline Methods} 
In this study, our primary focus lies in exploring LLM-based automated heuristic design approaches. Consequently, we compare the two closest related works, namely FunSearch~\citep{romera2024mathematical} and EoH~\citep{liu2024eoh}. The details can be found in Appendix~\ref{sup:baseline_details}.

\begin{figure*}[h!]
\centering
\subfloat[Pareto Front]{\includegraphics[width = 0.25\linewidth]{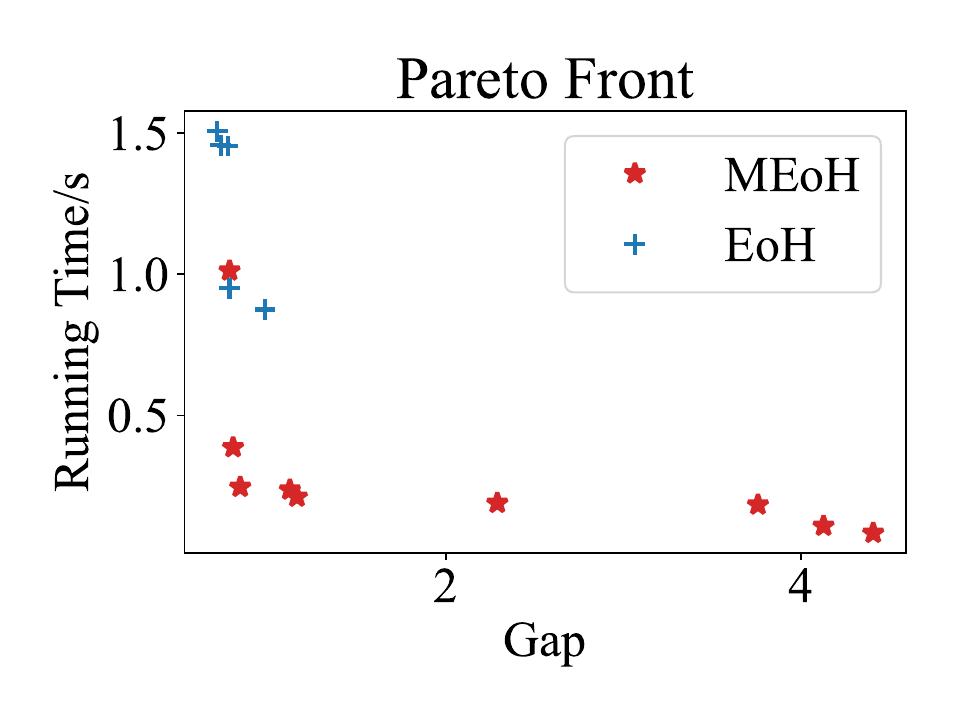}}
\subfloat[HV] {\includegraphics[width = 0.25\linewidth]{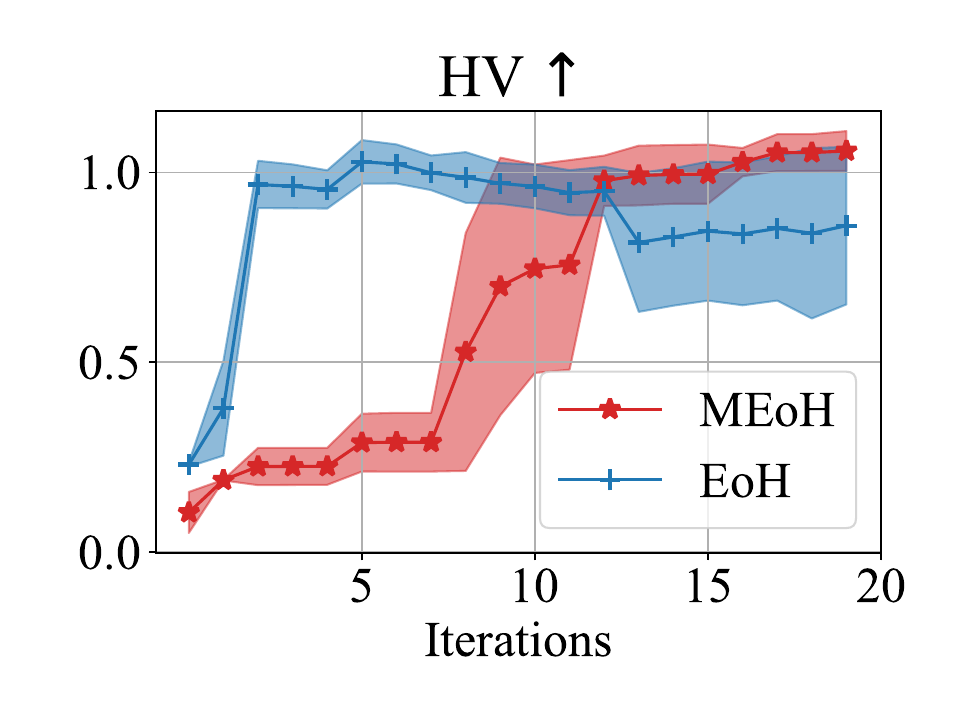}}
\subfloat[IGD]{\includegraphics[width = 0.25\linewidth]{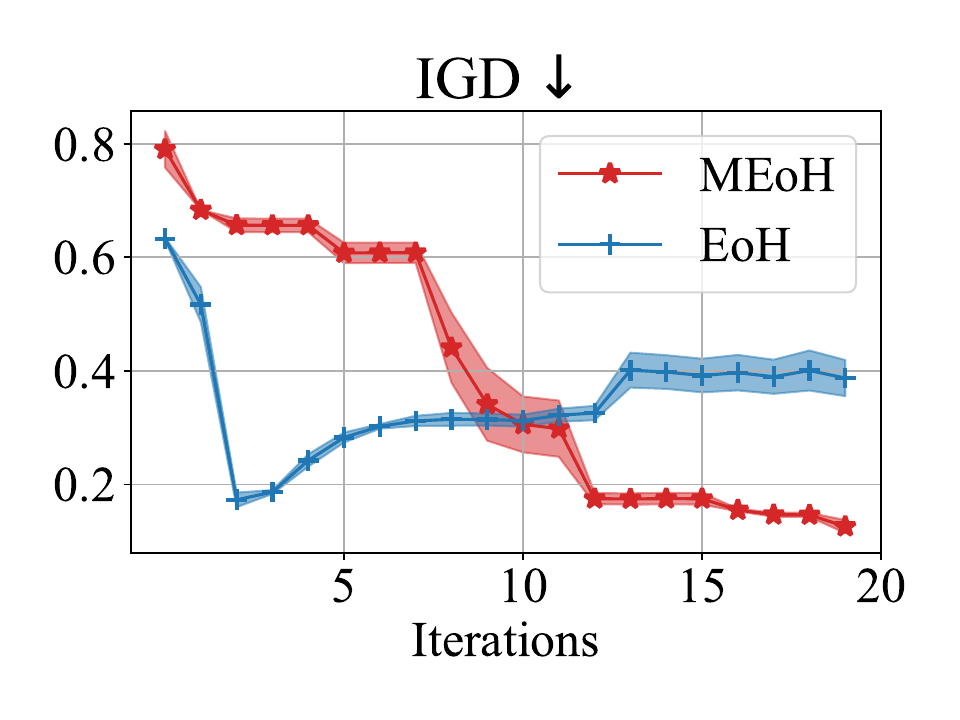}}
\subfloat[Score]{\includegraphics[width = 0.25\linewidth]{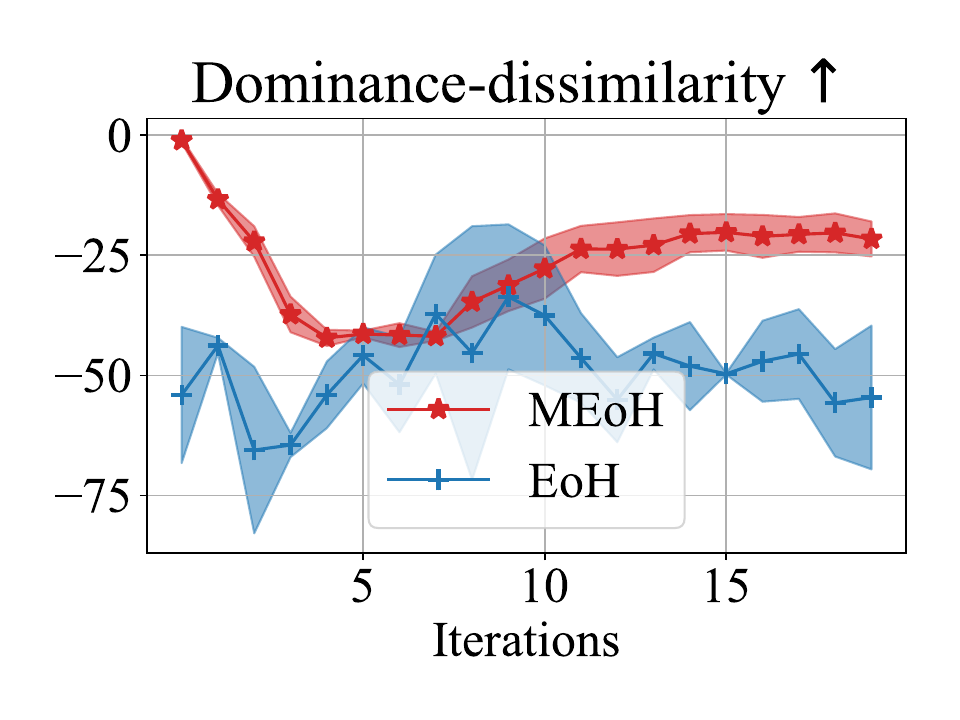}}
\caption{Comparations of EoH and MEoH on BPP5k.}
\label{fig: BPP}
\end{figure*}

\begin{figure*}[h!]
\centering
\subfloat[Pareto Front]{\includegraphics[width = 0.25\linewidth]{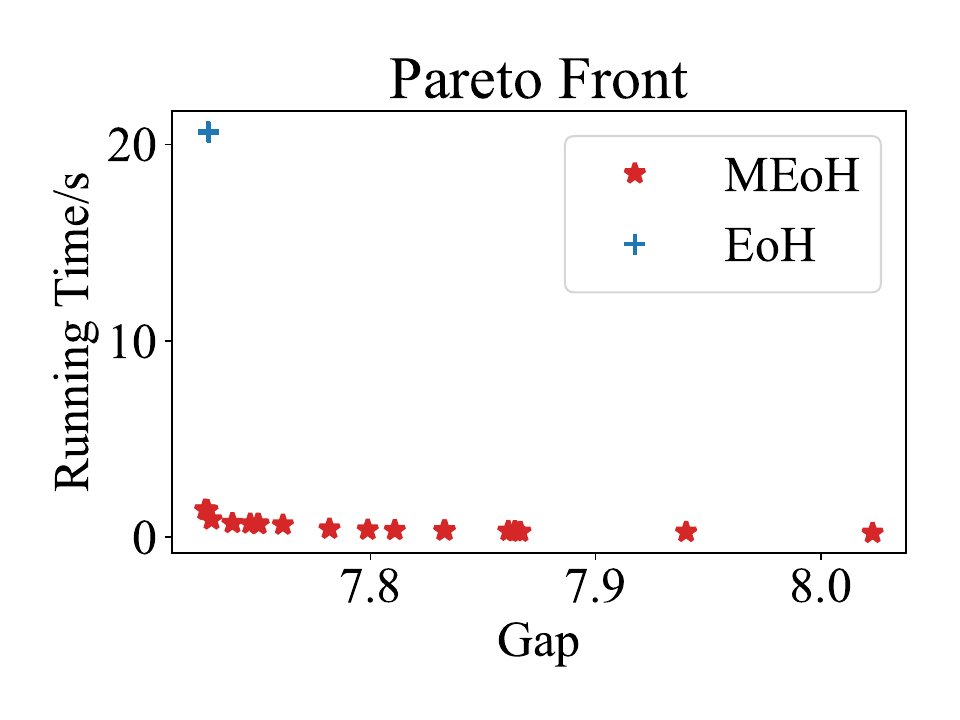}}
\subfloat[HV] {\includegraphics[width = 0.25\linewidth]{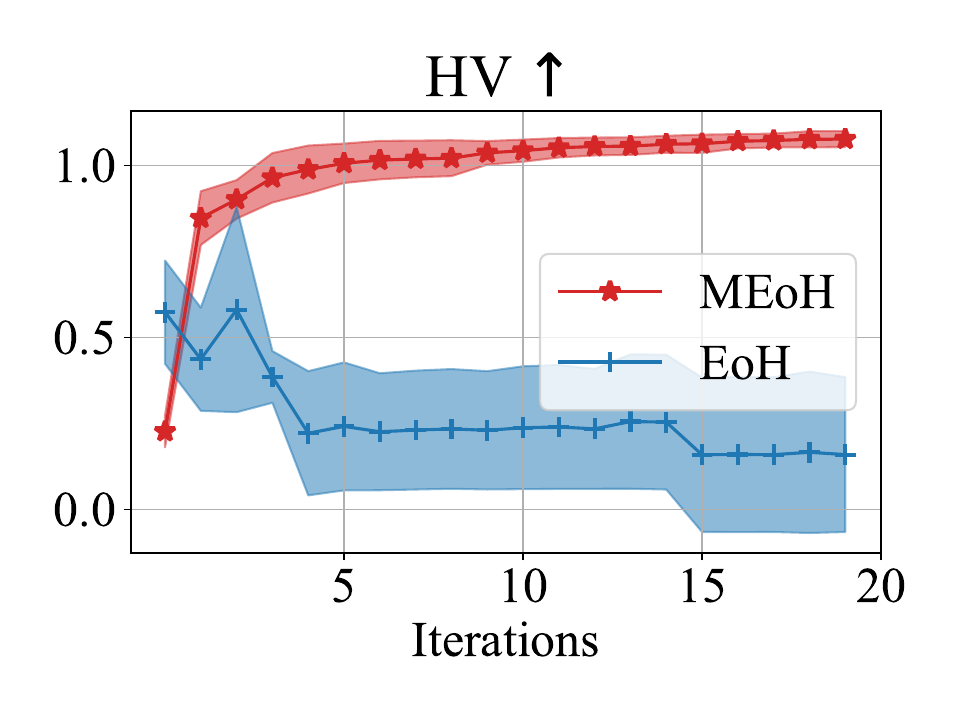}}
\subfloat[IGD]{\includegraphics[width = 0.25\linewidth]{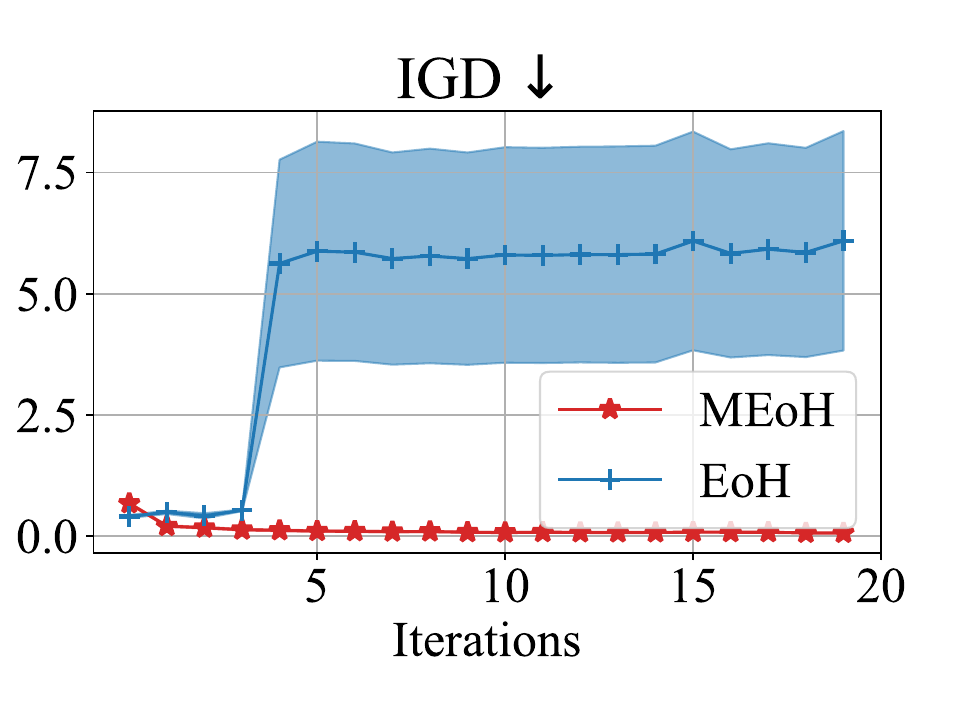}}
\subfloat[Score]{\includegraphics[width = 0.25\linewidth]{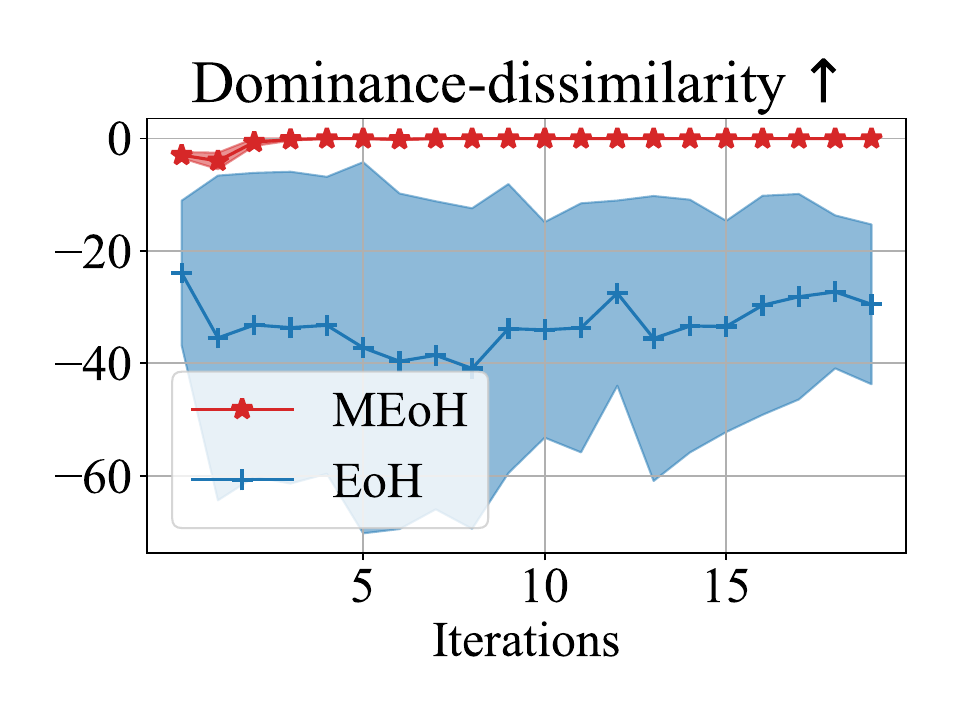}}
\caption{Comparations of EoH and MEoH on TSP100.}
\label{fig: TSP_GLS}
\end{figure*}

\subsection{Experimental Results}

\subsubsection{Convergence Analysis}

The curve of HV and IGD for the heuristic populations generated in each iteration on BPP are displayed in Figure~\ref{fig: BPP}(b) and (c), respectively. As EoH only pursues optimal gaps without considering diversity, the HV and IGD become worse as the evolution progresses. In contrast, MEoH systematically takes into account both the optimal gap and running time. As a result, MEoH achieves notably higher HV and lower IGD, indicating significantly better multi-objective trade-off results. Figure~\ref{fig: TSP_GLS}(b) and (c) provide more evidence on TSP. MEoH converges faster and clearly outperforms EoH in terms of HV and IGD. Additionally, the average dominance-dissimilarity score is shown in Figure~\ref{fig: BPP} (d) and Figure~\ref{fig: TSP_GLS} (d). Results demonstrate the superiority of MEoH and the efficiency of our dominance-dissimilarity mechanism in maintaining population diversity. The details can be found in Appendix~\ref{sup:DD_score}.

\subsubsection{Pareto Fronts}

Figure~\ref{fig: BPP}(a) and Figure~\ref{fig: TSP_GLS}(a) compare the non-dominated heuristics of the final population obtained by MEoH and EoH. Results show that 1) MEoH generates a diverse set of heuristics with different trade-offs over the two objectives. In contrast, EoH only finds similar heuristics that cover a much smaller region in the objective space. 2) The heuristics obtained from MEoH can significantly reduce the running time (up to 10 times) when achieving a similar optimal gap. 

\subsubsection{Performance Measurement}
1) \textit{BPP: } To comprehensively evaluate the performance of our MEoH in more general cases, we test FunSearch, EoH, and MEoH on various problem instances with different sizes and capacities. The problem sizes in our test include 5k, 10k, and 100k, and the capacities of the bins are set at $100$ and $500$. Each test set consists of five instances sampled from Weibull distribution~\citep{romera2024mathematical}. The average gap with reference to the relaxation lower bound $lb$ and the running time are shown in Table~\ref{tab: BPP}. For the in-distribution instances, i.e., the bin capacity is $100$, all of these three frameworks exhibit promising performance in terms of the optimal gap, and the running time of MEoH heuristics are significantly less than the counterparts of FunSearch and EoH, especially in large-size instances, i.e., BPP100k. MEoH heuristics achieve competitive performance compared to EoH but do so in significantly less running time (up to 10 times faster). In contrast, for out-distribution instances, i.e., the bin capacity is $500$, the performance of FunSearch heuristics drastically deteriorates in terms of the optimal gap. On the other hand, both EoH and MEoH heuristics exhibit promising performances in such scenarios. Notably, MEoH demonstrates a balanced trade-off between the optimal gap and running time, showcasing its effectiveness in handling out-distribution instances efficiently.

\begin{table}[h!]
\centering
\resizebox{\linewidth}{!}{
\begin{tabular}{l|rr|rr|rr}
\toprule
\multirow{2}{*}{Weibull} & \multicolumn{2}{c}{FunSearch} & \multicolumn{2}{c}{EoH} & \multicolumn{2}{c}{MEoH} \\
\cmidrule(lr){2-3}\cmidrule(lr){4-5}\cmidrule(lr){6-7}
& Gap& Time/s & Gap & Time/s & Gap & Time/s \\
\midrule
5k C100  & 0.802\%& 0.728& \cellcolor{lightgray} 0.753\%& 1.362  & 1.387\% & \cellcolor{lightgray} 0.191  \\
10k C100 & 2.595\%& 2.128& \cellcolor{lightgray} 0.537\%& 5.128  & 0.651\% & \cellcolor{lightgray} 0.650  \\
100k C100& 3.319\%& 195.734  & 0.391\%& 502.938& \cellcolor{lightgray} 0.080\% & \cellcolor{lightgray} 59.078 \\
\midrule
Avg. & 2.239\% & 66.197 & \cellcolor{lightgray}0.560\% & 169.809 & 0.706\% & \cellcolor{lightgray}19.973 \\
\midrule
5k C500  & 29.494\%   & 0.750& \cellcolor{lightgray}0.100\%& 1.672  & 0.351\% & \cellcolor{lightgray}0.100  \\
10k C500 & 47.734\%   & 2.459& \cellcolor{lightgray}0.125\%& 6.337  & 0.473\% & \cellcolor{lightgray}0.306  \\
100k C500& 53.640\%   & 259.094  & \cellcolor{lightgray}0.099\%& 646.828& 0.410\% & \cellcolor{lightgray}22.078 \\
\midrule
Avg. & 43.623\% & 87.434 & \cellcolor{lightgray} 0.108\% & 218.279 & 0.411\% & \cellcolor{lightgray} 7.495 \\
\bottomrule
\end{tabular}}
\caption{Results of in- and out-of-distribution BPP.}
\label{tab: BPP}
\end{table}

2) \textit{TSP: } 
We evaluate these three methods on randomly generated TSP instances comprising $100$, $500$, and $1,000$ nodes and a variety of TSP instances with up to $1,002$ nodes from TSPLIB~\citep{Rein91}. Table~\ref{tab: TSP_GLS_random} and Table~\ref{tab: TSP_GLS} display the gap compared to the best-known solution (for the randomly generated instances, the best-known solutions are obtained using the Concorde solver~\cite{applegate2006concorde}) and the corresponding running times. As shown in Table~\ref{tab: TSP_GLS_random}, FunSearch and MEoH (Best) heuristics exhibit promising performance on TSP100 and TSP500 instances. In general, MEoH provides a set of heuristics that enable trade-offs between optimality and efficiency. As shown in Table~\ref{tab: TSP_GLS}, for smaller instances (up to $200$ nodes), the MEoH heuristics demonstrate superior performance in terms of both the optimal gap and running time. For larger instances ($201$ to $1,002$ nodes), MEoH still outperforms in running time, although slightly lagging behind EoH in terms of the optimal gap. The details can be found in Appendix~\ref{sup:tsplib}.

\begin{table}[h!]
\centering
\resizebox{\linewidth}{!}{
\begin{tabular}{l|rr|rr|rr}
\toprule
           & \multicolumn{2}{c}{TSP100} & \multicolumn{2}{c}{TSP500} & \multicolumn{2}{c}{TSP1000} \\
           \cmidrule(lr){2-3}\cmidrule(lr){4-5}\cmidrule(lr){6-7}
           & Gap          & Time/s      & Gap          & Time/s      & Gap          & Time/s       \\
\midrule
FunSearch  & \cellcolor{lightgray}0.100\%      & 1.452       & \cellcolor{lightgray}1.525\%      & 27.598      & \cellcolor{lightgray}2.344\%      & 161.124      \\
EoH        & 0.113\%      & 22.434      & 1.750\%      & 43.541      & 2.524\%      & 262.603      \\
\midrule
MEoH(Best) & 0.109\%      & 1.373       & 1.733\%      & 30.945      & 4.208\%      & 26.844       \\
MEoH(Fast) & 3.690\%      & \cellcolor{lightgray}0.175       & 4.402\%      & \cellcolor{lightgray}3.306       & 4.536\%      & \cellcolor{lightgray}21.900       \\
\bottomrule
\end{tabular}}
\caption{Results of in- and out-of-distribution randomly generated TSP.}
\label{tab: TSP_GLS_random}
\end{table}

\begin{table}[h!]
\centering
\resizebox{\linewidth}{!}{
\begin{tabular}{l|rr|rr|rr}
\toprule
\multirow{2}{*}{TSPLIB} & \multicolumn{2}{c}{FunSearch} & \multicolumn{2}{c}{EoH} & \multicolumn{2}{c}{MEoH} \\
\cmidrule(lr){2-3}\cmidrule(lr){4-5}\cmidrule(lr){6-7}
   & Gap & Time/s& Gap & Time/s & Gap & Time/s\\
\midrule
Avg. (0-200)& 0.050\%   & 3.418 & 0.093\%& 25.917 & \cellcolor{lightgray}0.018\% & \cellcolor{lightgray}2.354  \\
\midrule
Avg. (201-1002)& 1.535\%   & 419.613   & \cellcolor{lightgray}1.376\%& 1515.992   & 1.50\%  & \cellcolor{lightgray}355.754\\
\bottomrule
\end{tabular}}
\caption{Results of small and large TSPLIB instances.}
\label{tab: TSP_GLS}
\end{table}

\begin{table}[h!]
\scriptsize
\centering
\begin{tabular}{l|rr|rr}
\toprule
& \multicolumn{2}{c}{TSPLIB} & \multicolumn{2}{c}{BPP C100}  \\
\cmidrule(lr){2-3}\cmidrule(lr){4-5}
& Gap  & Time/s  & Gap   & Time/s   \\
\midrule
FunSearch   & 0.050\%  & 3.418   & 2.239\%   & 66.197   \\
EoH & 0.093\%  & 25.917  & \cellcolor{lightgray}0.560\%   & 169.809  \\
\midrule
MEoH (Best) & \cellcolor{lightgray}0.018\%  & 2.354   & 0.706\%   & 19.973   \\
MEoH (Fast) & 3.563\%  & \cellcolor{lightgray}0.138   & 4.326\%   & \cellcolor{lightgray}6.533\\
\bottomrule
\end{tabular}
\caption{Two top heuristics designed by MEoH.}
\label{tab: MEoH}
\end{table}

\subsection{Comparison to Conventional MOEAs} 

\begin{figure}[h!]
\centering
\subfloat[HV] {\includegraphics[width = 0.5\linewidth]{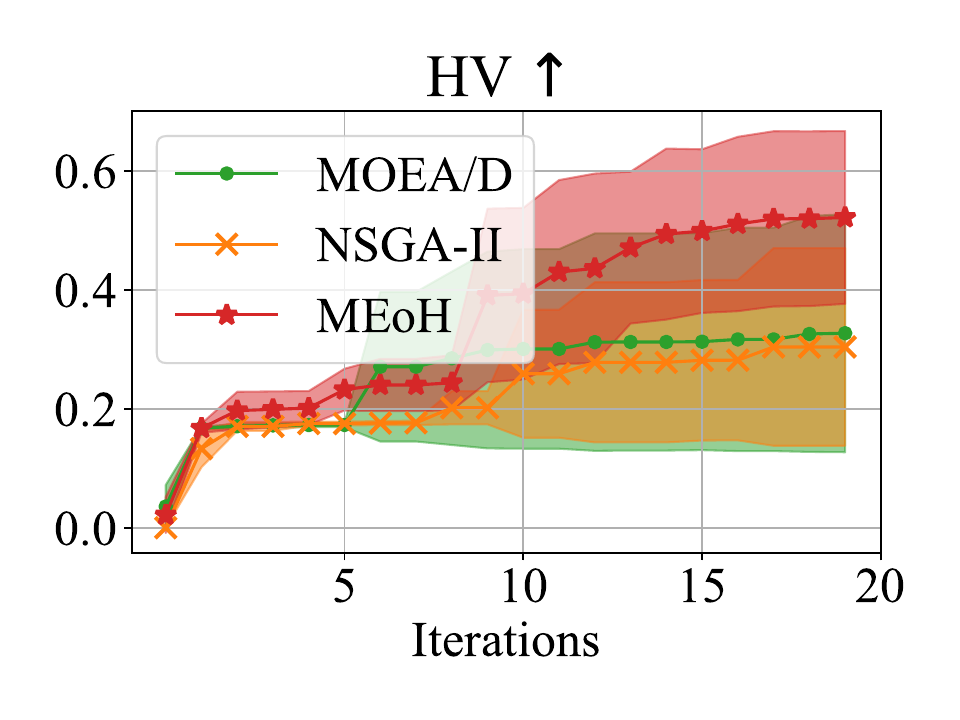}}
\subfloat[IGD]{\includegraphics[width = 0.5\linewidth]{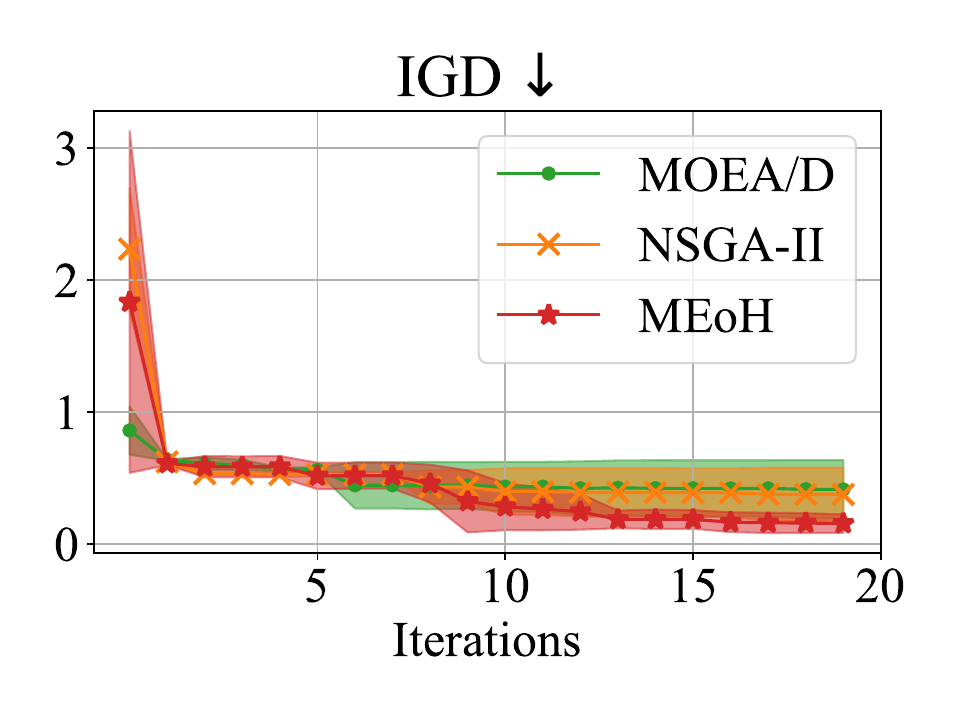}}
\caption{Comparison to conventional MOEAs on BPP.}
\label{fig: BPP_ablation}
\end{figure}

\begin{figure}[h!]
\centering
\subfloat[HV] {\includegraphics[width = 0.5\linewidth]{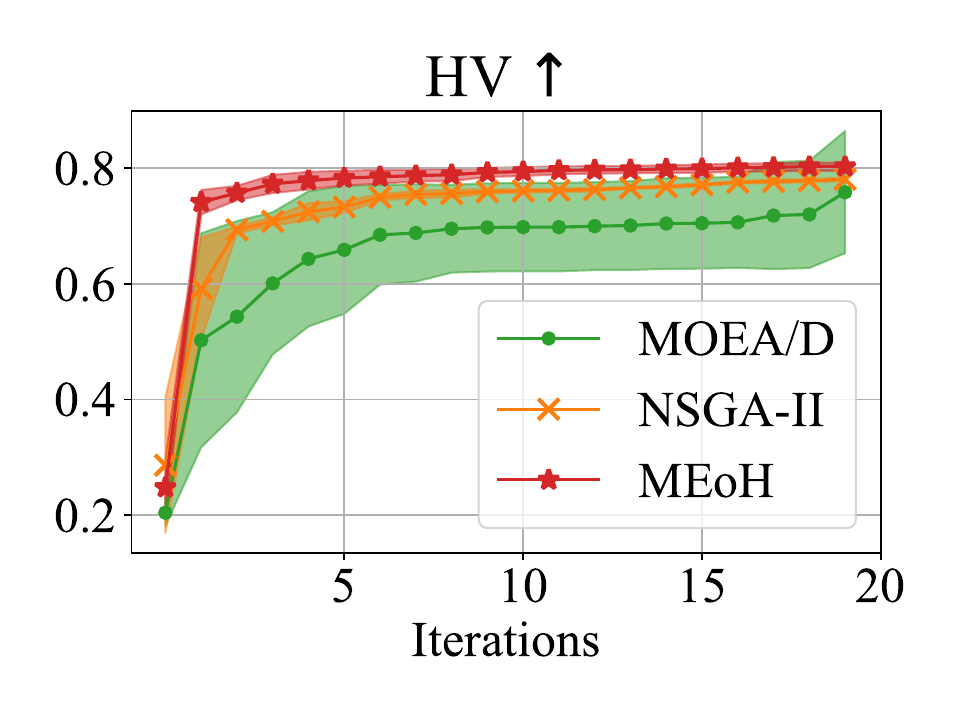}}
\subfloat[IGD]{\includegraphics[width = 0.5\linewidth]{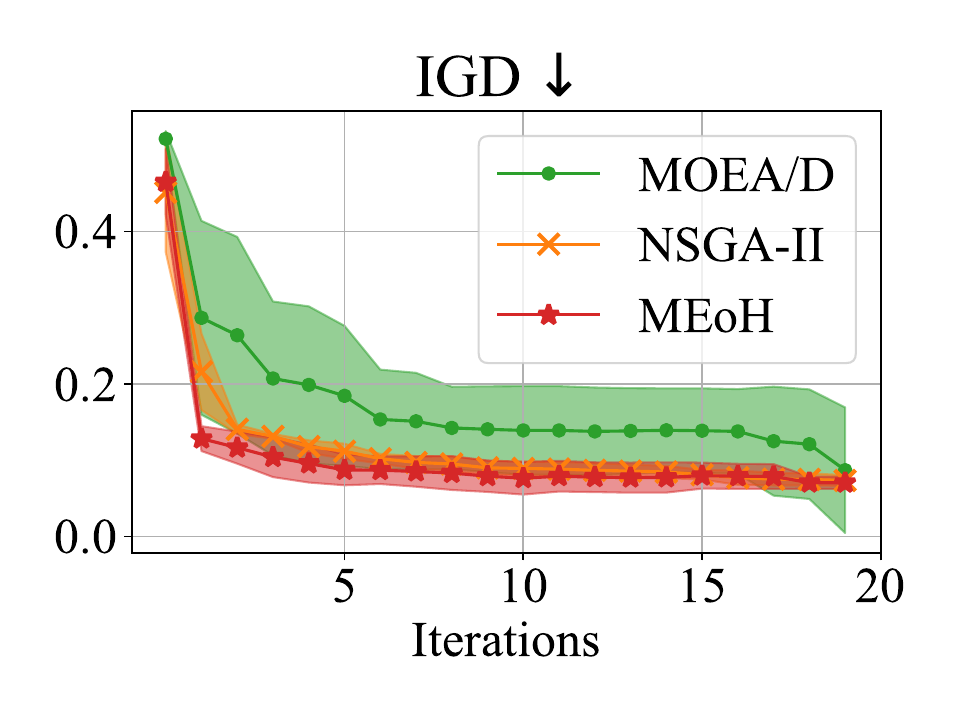}}
\caption{Comparison to conventional MOEAs on TSP.}
\label{fig: TSP_ablation}
\end{figure}

In this section, we evaluate the impact of our proposed dominance-dissimilarity mechanism on the optimization process and compare to two representative MOEAs: NSGA-II~\cite{deb2002fast} and MOEA/D~\cite{zhang2007moea}. 

Figure~\ref{fig: BPP_ablation} and Figure~\ref{fig: TSP_ablation} depict the results on BPP and TSP, respectively. MEoH can obtain the best HV and IGD. Our findings highlight the effectiveness of our dominance-dissimilarity mechanism, which integrates considerations from both the search and objective spaces, in improving the optimization process. 

\begin{figure}[th]
\centering
{\includegraphics[width = 0.6\linewidth]{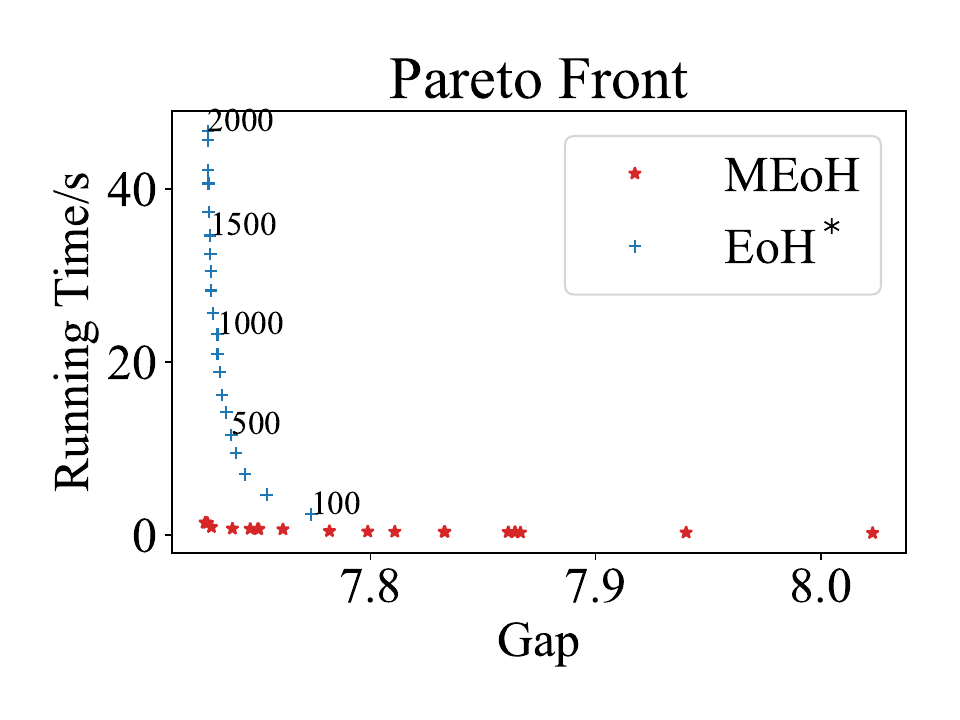}}
\caption{Comparations of the non-dominated heuristics generated by MEoH and the any-time performance of the best heuristic generated by EoH (termed as $\text{EoH}^*$) on TSP.}
\label{fig: TSP_GLS_best_EoH}
\end{figure}

\subsection{Comparison to Any-time Performance} 

The performance of a single heuristic at any given time can provide a set of heuristics that offer different trade-offs between optimal gap and running time. For instance, reducing the number of iterations in GLS from $1,000$ to $100$ results in a decrease in running time but a deterioration in the optimal gap. By comparing the heuristics generated by MEoH to the best heuristic produced by EoH, we can further illustrate the benefits of multi-objective heuristic design. We evaluate the performance of the best EoH heuristic with varying numbers of iterations. Figure~\ref{fig: TSP_GLS_best_EoH} demonstrates that the heuristics generated by MEoH outperform those of EoH. Even the best EoH heuristic with 100 iterations falls short in terms of running time and optimal gap compared to all MEoH heuristics. Additionally, while the best EoH heuristic with $2,000$ iterations can achieve competitive optimality, it lags behind in running time by approximately 20 times. 

\section{Conclusion, Limitation, and Future Work}

\paragraph{Conclusion}
This paper develops a novel framework, termed MEoH, for LLM-based multi-objective automatic heuristic design. We propose a dominance-dissimilarity mechanism for effective search in the discrete and complex heuristic space. We demonstrate MEoH on two widely-studied combinatorial optimization problems to optimize both heuristics' optimal gap and running time. Results show that MEoH significantly outperforms existing LLM-based heuristic design methods including FunSearch and EoH in producing trade-off heuristics over multiple objectives. The efficiency can be increased dramatically up to 10 times with a close optimal gap. Moreover, additional ablation studies and visualization of the evolution process validate the superiority of MEoH over conventional MOEAs and the effectiveness of the proposed dominance-dissimilarity mechanism in multi-objective automatic heuristic design. 

\paragraph{Limitation and Future Work} Although we have demonstrated the effectiveness of MEoH primarily on two objectives, and three objectives in Appendix~\ref{sup:3o}, we aim to investigate the performance of MEoH on many-objective cases and a broader range of heuristic design tasks.

\section*{Acknowledgments}
This work was supported by the Research Grants Council of the Hong Kong Special Administrative Region, China (GRF Project No. CityU 11215622), the National Natural Science Foundation of China (Grant No. 62106096 and Grant No. 62476118), the Natural Science Foundation of Guangdong Province (Grant No. 2024A1515011759), the National Natural Science Foundation of Shenzhen (Grant No. JCYJ20220530113013031).

\bibliography{aaai25}

\clearpage
\newpage
\appendix
\onecolumn

\section{Algorithm Details}~\label{sup:algorithms}

In this part, we elaborate on the details of parent selection and population management used in our proposed MEoH, as shown in Algorithm~\ref{alg:ParentsSelection} and Algorithm~\ref{alg:Pop_Management}, respectively. 

\paragraph{Calculation of Dominance-dissimilarity Score} The lines 3-16 in Algorithm~\ref{alg:ParentsSelection} and the lines 4-17 in Algorithm~\ref{alg:Pop_Management} are almost identical, illustrating the computation of the dominance-dissimilarity score. Specifically, two square matrices, namely the dissimilarity score matrix $\boldsymbol{S}$ and the dominance mask matrix $\boldsymbol{D}$, are initialized to be zeros. Each heuristic within the population is compared in pairs, with their dissimilarity (negative AST similarity) and dominance relationships recorded in the corresponding matrices. Subsequently, these matrices are element-wise multiplied to yield the dominance-dissimilarity score matrix $\boldsymbol{S}^\prime$. The dominance-dissimilarity vector $\boldsymbol{v}$ is then derived by summing the columns of $\boldsymbol{S}^\prime$. This vector encapsulates a blend of dominance and dissimilarity considerations, guiding the following parent selection and population management. 

\paragraph{Parent Selection} For parent selection, as delineated in Algorithm~\ref{alg:ParentsSelection}, the dominance-dissimilarity vector $\boldsymbol{v}$ is leveraged to construct a probability distribution $\boldsymbol{\pi}$ using the softmax function. The parents are subsequently sampled based on this distribution to strike a balance between exploration and exploitation. 

\paragraph{Population Management} For population management, as shown in Algorithm~\ref{alg:Pop_Management}, the dominance-dissimilarity vector $\boldsymbol{v}$ is descending sorted, and the resulting indices $\boldsymbol{k}$ are utilized to truncate the population, and the first $N$ individuals consists the new population $\boldsymbol{P}^\prime$.

\begin{algorithm}[htbp]
\setcounter{algorithm}{0}  
\begin{algorithmic}[1]
\STATE \textbf{Input:} Population $\boldsymbol{P}$; Population size $N$; Parent selection size $d$.
\STATE \textbf{Output:} Selected parents $\boldsymbol{P}_{parent}$.\\
\STATE Initialize the dissimilarity score matrix $\boldsymbol{S}$ as an $N \times N$ matrix filled with zeros;
\STATE Initialize the dominance mask matrix $\boldsymbol{D}$ as an $N \times N$ matrix filled with zeros;
\FOR{$i = 1,\ldots,N$}
\FOR{$j = 1,\ldots,N$}
\IF{$i \neq j$}
\STATE $\boldsymbol{S}[i, j] \leftarrow - $AST($\boldsymbol{P}[i], \boldsymbol{P}[j]$);
\IF{$\boldsymbol{P}[i] \prec \boldsymbol{P}[j]$}
\STATE $\boldsymbol{D}[i, j] \leftarrow 1$;
\ENDIF
\ENDIF
\ENDFOR
\ENDFOR
\STATE $\boldsymbol{S}^\prime \leftarrow \boldsymbol{S} \odot \boldsymbol{D}$ 
\STATE $\boldsymbol{v} \leftarrow$ ColumnwiseSum($\boldsymbol{S}^\prime$)
\STATE $\boldsymbol{\pi} \leftarrow$ Softmax($\boldsymbol{v}$)
\STATE $\boldsymbol{P}_{parent} \leftarrow$ Sample($\boldsymbol{P}, \boldsymbol{\pi}, d$)
\end{algorithmic}
\caption{ParentSelection}
\label{alg:ParentsSelection}
\end{algorithm}

\begin{algorithm}[htbp]
\begin{algorithmic}[1]
\STATE \textbf{Input:} Population $\boldsymbol{P}$; Population size $N$.
\STATE \textbf{Output:} New population $\boldsymbol{P}^\prime$.\\
\STATE Current population size $N^\prime \leftarrow$ size($\boldsymbol{P}$)
\STATE Initialize the dissimilarity score matrix $\boldsymbol{S}$ as an $N^\prime \times N^\prime$ matrix filled with zeros;
\STATE Initialize the dominance mask matrix $\boldsymbol{D}$ as an $N^\prime \times N^\prime$ matrix filled with zeros;
\FOR{$i = 1,\ldots,N$}
\FOR{$j = 1,\ldots,N$}
\IF{$i \neq j$}
\STATE $\boldsymbol{S}[i, j] \leftarrow - $AST($\boldsymbol{P}[i], \boldsymbol{P}[j]$);
\IF{$\boldsymbol{P}[i] \prec \boldsymbol{P}[j]$}
\STATE $\boldsymbol{D}[i, j] \leftarrow 1$;
\ENDIF
\ENDIF
\ENDFOR
\ENDFOR
\STATE $\boldsymbol{S}^\prime \leftarrow \boldsymbol{S} \odot \boldsymbol{D}$ 
\STATE $\boldsymbol{v} \leftarrow$ ColumnwiseSum($\boldsymbol{S}^\prime$)
\STATE $\boldsymbol{k} \leftarrow$ DescendingSortedIndexes($\boldsymbol{v}$)
\STATE Initialize a new population $\boldsymbol{P}^\prime \leftarrow \emptyset$
\FOR{$i = 1,\ldots,N$}
\STATE $\boldsymbol{P}^\prime \leftarrow \boldsymbol{P}^\prime \cup \boldsymbol{P}[\boldsymbol{k}[i]]$
\ENDFOR
\end{algorithmic}
\caption{PopulationManagement}
\label{alg:Pop_Management}
\end{algorithm}

\newpage
~\newpage

\section{Heuristic Design Task Details}~\label{sup:problem_details}

We demonstrate the proposed method on two heuristic design tasks: 1) heuristics design for online Bin Packing Problem (BPP) and 2) heuristic design for guided local search for Traveling Salesman Problem (TSP). We introduce the detailed heuristic design settings for each task.

\subsection{BPP}
In online Bin Packing Problem (BPP)~\citep{seiden2002online}, a set of items, each with its own weight, needs to be packed into bins with a predetermined capacity. The objective of the BPP is to minimize the total number of bins required to accommodate all the items. In an online scenario, items are packed as they are received without prior knowledge.

The heuristic operates by loading items sequentially in an online fashion, requiring only the selection of the best bin at each iteration. This designed function scores bins based on their remaining capacities and the size of the arriving item, with the highest scoring bin chosen for each iteration. The function takes two inputs - the size of the arriving item and the remaining capacities of the bins - and outputs a vector that ranks the bins accordingly. A task description used in the prompt and the Python code snippet template are illustrated as follows:

\begin{figure}[H]
\setcounter{figure}{0}  
\centering
\Ovalbox{
\begin{minipage}{0.9\textwidth}
\vspace{10pt}
\begin{adjustwidth}{5pt}{}
\textcolor{Blue}{\textbf{Task Description: }Implement a function that returns the priority with which we want to add an item to each bin. }

\vspace{10pt}
\textcolor{Blue}{\textbf{Template Program: }}

\begin{minted}[breaklines]{python}
import numpy as np

def priority(item: float, bins: np.ndarray) -> np.ndarray:
    """Returns priority with which we want to add item to each bin.
    Args:
        item: Size of item to be added to the bin.
        bins: Array of capacities for each bin.
    Return:
        Array of same size as bins with priority score of each bin.
    """
    return item - bins
\end{minted}

\vspace{10pt}
\end{adjustwidth}
\end{minipage}
}
\caption{BPP heuristic design description and template program.}
\label{fig: BPP_prompt}
\end{figure}

\subsection{TSP}

For TSP, one of the widely used metaheuristics, Guided Local Search (GLS), is used~\citep{voudouris2010guided}. The pipeline of GLS is as follows:

\textbf{Step 1:} Create an initial solution using nearest neighbor constructive heuristics.

\textbf{Step 2:} Local Search Stage: Perform a local search (swap and relocate) to improve the current solution and generate a local optimal solution.

\textbf{Step 3:} Perturbation Stage: Update the distance matrix. Perform another local search based on the updated distance matrix to perturb the local optimal solution to escape from local optimality.

\textbf{Steps 2} and \textbf{3} are iteratively repeated until the stopping criterion (maximum number of iterations set to $1,000$ in the experiments) is satisfied. The best solution obtained throughout the iterations is considered the final solution.

Our goal is to develop a heuristic to update the distance matrix in the perturbation step. The task description provided in the prompt and the template of the Python code snippet is outlined below. The inputs include the original distance matrix, the local optimal solution, and the frequency of edge usage in perturbation. The output should be the updated distance matrix.

\begin{figure}[H]
\centering
\Ovalbox{
\begin{minipage}{0.9\textwidth}
\vspace{10pt}
\begin{adjustwidth}{5pt}{}
\textcolor{Blue}{\textbf{Task Description: }Given an edge distance matrix and a local optimal route, please help me design a strategy to update the distance matrix to avoid being trapped in the local optimum with the final goal of finding a tour with minimized distance. You should create a heuristic for me to update the edge distance matrix. }

\vspace{10pt}

\textcolor{Blue}{\textbf{Template Program: }}

\begin{minted}[breaklines]{python}
import numpy as np
def update_edge_distance(edge_distance: np.ndarray, local_opt_tour: np.ndarray, edge_n_used: np.ndarray) -> np.ndarray:
    """
    Design a novel algorithm to update the distance matrix.

    Args:
    edge_distance: A matrix of the distance.
    local_opt_tour: An array of the local optimal tour of IDs.
    edge_n_used: A matrix of the number of each edge used during permutation.

    Return:
    updated_edge_distance: A matrix of the updated distance.
    """
    updated_edge_distance = np.copy(edge_distance)

    # Calculate combined importance and frequency factor
    updated_edge_distance = edge_distance

    return updated_edge_distance
\end{minted}

\vspace{10pt}
\end{adjustwidth}
\end{minipage}
}
\caption{TSP heuristic design task description and template program.}
\label{fig: TSP_prompt}
\end{figure}

\section{Baseline Settings}~\label{sup:baseline_details}

In this work, we employ FunSearch~\cite{romera2024mathematical} and EoH~\cite{liu2024eoh} as baseline. For EoH, we inherit the default settings, including the number of iterations $T=20$, the parent selection size $d=5$, and the population size $N=10$ for the TSP and $N=20$ for the BPP. Our MEoH also follows these settings. In summary,  $1,000$ heuristics are generated for solving TSP, and $2,000$ heuristics for BPP. For FunSearch, we also adopt the default settings, the number of islands is $10$ and the number of samples for each prompt is $4$. FunSearch generates $10,000$ heuristics for solving BPP and TSP.

\section{Metric Definition}~\label{sup:metric_definition}

\subsection{HV}
\label{sup: HV}
Hypervolume (HV) is calculated as follows:
\begin{equation}
{\rm HV}(\mathcal{P}, \boldsymbol{r}^*) = {\rm VOL}\Big(\mathop{\cup}\limits_{\boldsymbol{v}\in \mathcal{P}}[v_1,r_1^*]\times \ldots \times [v_m,r_m^*]\Big),
\end{equation}
where $\mathcal{P}$ represents the approximate Pareto front obtained by an automated heuristic design approach, $\boldsymbol{v}=(v_1,\ldots,v_m)^\intercal$ denotes the corresponding objective vector, ${\rm VOL}(\cdot)$ represents the Lebesgue measure, and $\boldsymbol{r}^* = (r_1^*,\ldots ,r_m^*)^\intercal$ is a reference objective vector.

To account for variations in HV values across different objective domains, i.e., the scalar of intrinsic objective value and the running time, we normalized each objective value for each instance. Specifically, the generated heuristic $\boldsymbol{x}$ can be normalized in the objective space using the approximated ideal point $\boldsymbol{z}^\text{ideal} = (z_1^\text{ideal}, \ldots, z_M^\text{ideal})^\intercal$ and the approximated nadir point $\boldsymbol{z}^{\text{nadir}} = (z_1^{\text{nadir}}, \ldots, z_M^{\text{nadir}})^\intercal$ derived from the union of all approximated Pareto-front $\mathcal{P}$ as 
\begin{equation}
f_i^\prime(\boldsymbol{x})=\frac{f_i(\boldsymbol{x})-z_i^\text{ideal}}{z_i^{\text{nadir}}-z_i^\text{ideal}},
\end{equation}
where $z_i^\text{ideal} = \min\{v_i | \boldsymbol{v} \in \mathcal{P}\}$ and $z_i^\text{nadir} = \max\{v_i | \boldsymbol{v} \in \mathcal{P}\}$, $\forall i \in \{1, \ldots, M\}$. Consequently, the value of each objective is normalized to $[0, 1]$. Based on that, the reference point $\boldsymbol{r}^*=(1.1,\ldots,1.1)^\intercal$.

\subsection{IGD} 
\label{sup: IGD}
Inverted Generational Distance (IGD) measures the convergence and diversity of the obtained Pareto front approximation concerning the true Pareto front. It is calculated as follows:
\begin{equation}
\text{IGD}(P, P^*) = \frac{1}{{\lvert P^* \rvert}} \sum_{p \in P^*} \min_{q \in P} d(p, q),
\end{equation}
where $P$ is the set of decision vectors, i.e, the approximated Pareto front. $P^*$ is the true Pareto front, $\lvert P^* \rvert$ is the number of points in the true Pareto front
$d(p, q)$ is the Euclidean distance between the points $p$ and $q$ in the objective space. 

The IGD calculates the average distance from the true Pareto front points to their nearest neighbor in the approximated Pareto front. A lower IGD value indicates a better approximation of the true Pareto front.

It's important to note that the true Pareto front is required for calculating the IGD metric, which may not always be available in many cases. So, a reference set of well-distributed Pareto-optimal heuristics is often used as an approximation of the true Pareto front, here the reference set is the nondominated set derived from the union of all generated heuristics.

\section{TSPLIB Results}~\label{sup:tsplib}

\begin{table}[H]
\centering
\caption{Results of small and large TSPLIB instances.}
\label{tab: TSP_GLS_details}
\resizebox{0.5\linewidth}{!}{
\begin{tabular}{l|rr|rr|rr}
\toprule
\multirow{2}{*}{TSPLIB} & \multicolumn{2}{c}{FunSearch} & \multicolumn{2}{c}{EoH} & \multicolumn{2}{c}{MEoH} \\
\cmidrule(lr){2-3}\cmidrule(lr){4-5}\cmidrule(lr){6-7}
   & Gap & Time/s& Gap & Time/s & Gap & Time/s\\
\midrule
berlin52& 0.000\%   & 0.484 & 0.000\%& 8.500  & 0.000\% & \cellcolor{lightgray}0.344  \\
ch130   & \cellcolor{lightgray} 0.156\%   & 2.031 & 0.233\%& 42.360 & 0.233\% & \cellcolor{lightgray}\cellcolor{lightgray}1.016  \\
ch150   & 0.306\%   & 2.500 & 0.502\%& 56.062 & \cellcolor{lightgray}0.000\% & \cellcolor{lightgray}1.250  \\
eil101  & \cellcolor{lightgray}0.000\%   & 28.391& 0.373\%& 56.031 & \cellcolor{lightgray}0.000\% & 2\cellcolor{lightgray}2.297 \\
eil51   & 0.000\%   & 0.515 & 0.000\%& 7.109  & 0.000\% & \cellcolor{lightgray}0.312  \\
eil76   & 0.183\%   & 0.938 & 0.107\%& 14.844 & \cellcolor{lightgray}0.000\% & \cellcolor{lightgray}0.531  \\
kroA100 & 0.000\%   & 1.407 & 0.000\%& 24.437 & 0.000\% & \cellcolor{lightgray}0.734  \\
kroC100 & 0.000\%   & 1.500 & 0.000\%& 24.781 & 0.000\% & \cellcolor{lightgray}0.703  \\
kroD100 & 0.000\%   & 1.578 & 0.000\%& 24.563 & 0.000\% & \cellcolor{lightgray}0.734  \\
lin105  & 0.000\%   & 1.812 & 0.000\%& 26.703 & 0.000\% & \cellcolor{lightgray}0.890  \\
pr76& 0.000\%   & 0.969 & 0.000\%& 14.469 & 0.000\% & \cellcolor{lightgray}0.546  \\
rd100   & 0.000\%   & 1.453 & 0.000\%& 24.266 & 0.000\% & \cellcolor{lightgray}0.750  \\
st70& 0.000\%   & 0.859 & 0.000\%& 12.797 & 0.000\% & \cellcolor{lightgray}0.500  \\
\midrule
Avg.& 0.050\%   & 3.418 & 0.093\%& 25.917 & \cellcolor{lightgray}0.018\% & \cellcolor{lightgray}2.354  \\
\midrule
a280& 0.195\%   & 378.656   & \cellcolor{lightgray}0.059\%& 640.453& 1.245\% & \cellcolor{lightgray}356.468\\
pcb442  & 1.389\%   & 932.093   & 1.714\%& 1694.547   & \cellcolor{lightgray}1.284\% & \cellcolor{lightgray}916.219\\
pr1002  & 2.878\%   & 354.813   & \cellcolor{lightgray}2.487\%& 3592.891   & 3.272\% & \cellcolor{lightgray}142.000\\
tsp225  & 1.679\%   & 12.891& 1.243\%& 136.078& \cellcolor{lightgray}0.197\% & \cellcolor{lightgray}8.328  \\
\midrule
Avg.& 1.535\%   & 419.613   & \cellcolor{lightgray}1.376\%& 1515.992   & 1.50\%  & \cellcolor{lightgray}355.754\\
\bottomrule
\end{tabular}}
\end{table}

\section{Visualization of Dominance-dissimilarity Scores}~\label{sup:DD_score}

\begin{figure}[H]
\centering
\subfloat[BPP, MEoH]{\includegraphics[width = 0.5\linewidth]{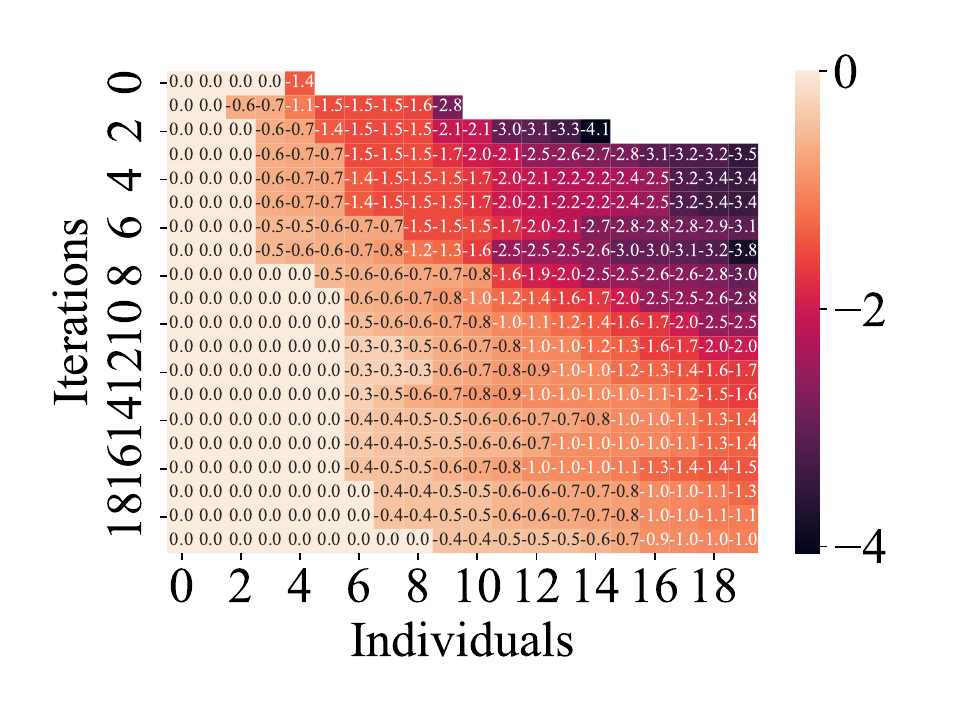}}
\subfloat[BPP, EoH]{\includegraphics[width = 0.5\linewidth]{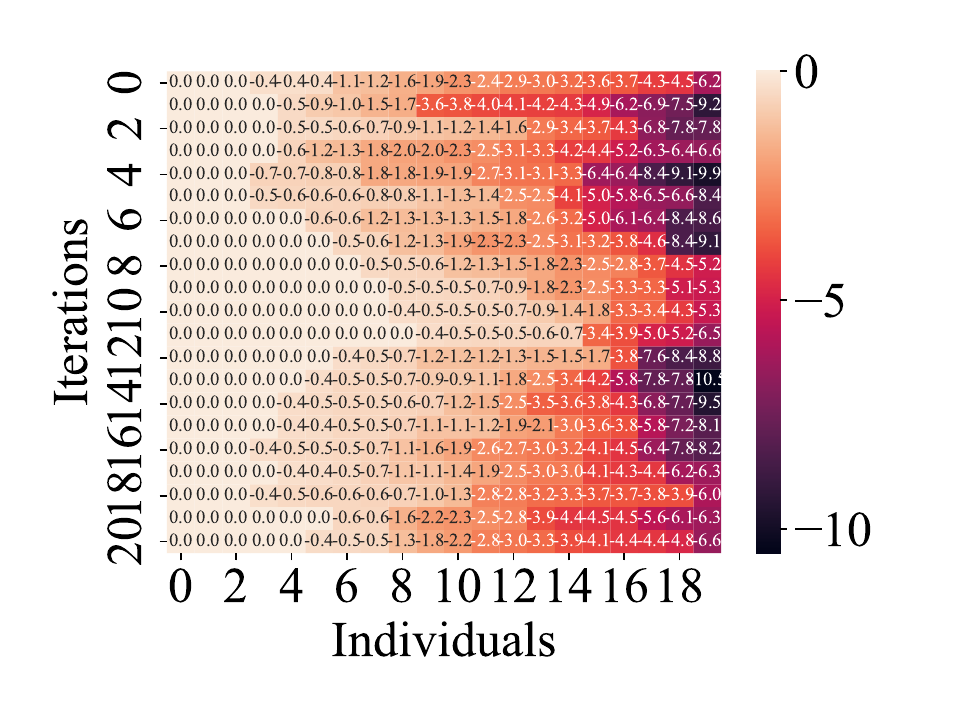}}
\newline
\subfloat[TSP, MEoH]{\includegraphics[width = 0.5\linewidth]{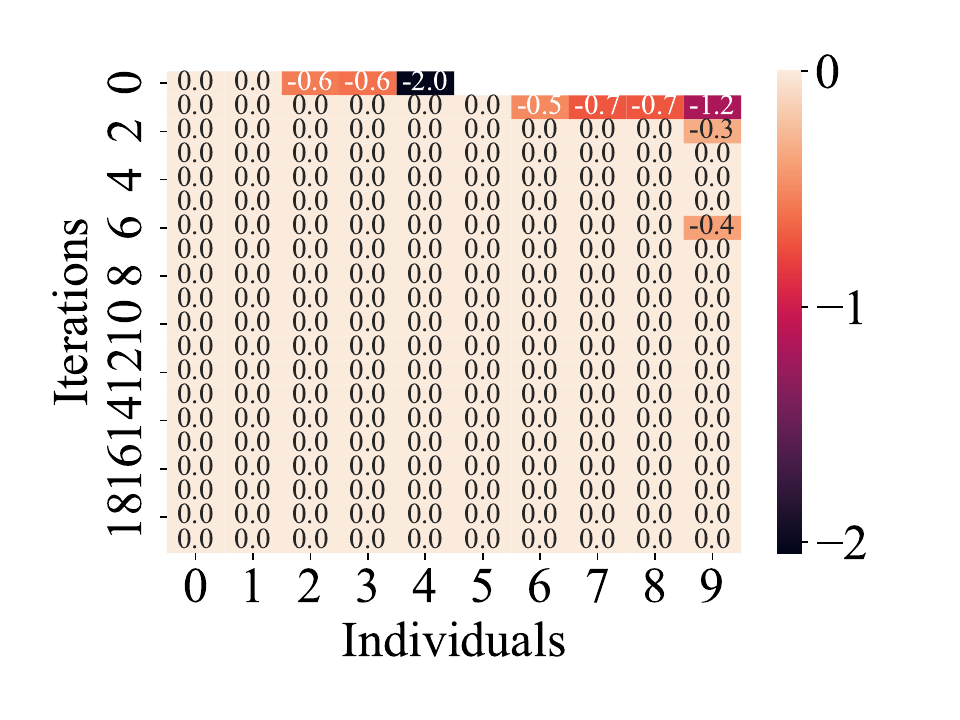}}
\subfloat[TSP, EoH]{\includegraphics[width = 0.5\linewidth]{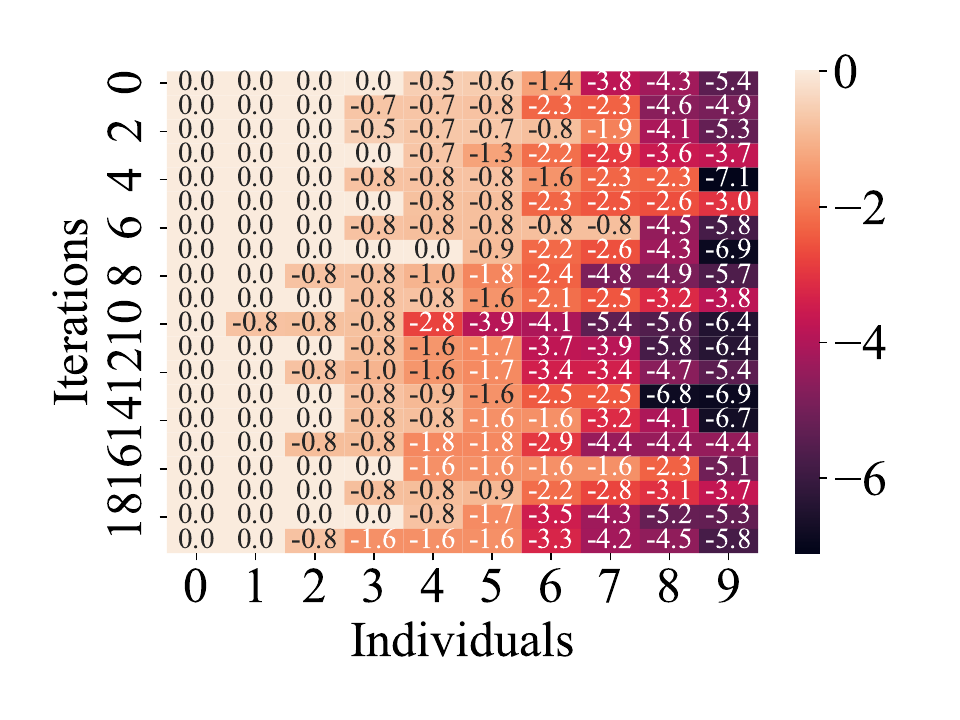}}
\caption{Visualization of the evolution of dominance-dissimilarity score.}
\label{fig:DD}
\end{figure}

We visualize the evolution of Dominance-dissimilarity Scores in Figure~\ref{fig:DD}. The x-axis is the heuristic index, and the y-axis is the iteration index. It is important to note that the presence of blank blocks in the early iterations indicates cases where the population is not filled, due to the generation of illegal code segments by LLM. As shown in Figure~\ref{fig:DD}, MEoH heuristics can maintain diversity during the evolutionary process, while the diversity of EoH drastically deteriorates. 

\section{Search Operators}~\label{sup:operator}

MEoH inherits $5$ search operators from EoH~\cite{liu2024eoh}. These operators are all implemented based on LLMs. In this part, the corresponding prompts will be elaborated. Generally, the prompt consists of operator-specific guidance, task description, and program template. For brevity, the task description and the program template are denoted as \textcolor{Blue}{\underline{\textbf{\$Task Description}}} and \textcolor{Blue}{\underline{\textbf{\$Program Template}}}, respectively.

\subsection{E1 Operator}
As shown in Figure~\ref{fig: E1_prompt}, the E1 operator is used to explore a new heuristic different from the $5$ selected heuristics. For simplicity, the heuristics including corresponding heuristic description and code are omitted.

\begin{figure}[H]
\centering
\Ovalbox{
\begin{minipage}{0.9\textwidth}
\vspace{10pt}

\begin{adjustwidth}{5pt}{}

\textcolor{Blue}{\underline{\textbf{\$Task Description}}}

\vspace{10pt}

\textcolor{Red}{I have $5$ existing algorithms with their codes as follows: }

\textcolor{Red}{$<$Algorithm description$>$: } ...

\textcolor{Red}{$<$Code$>$: } ... 

\vspace{10pt}

...

\vspace{10pt}

\textcolor{Red}{Please help me create a new algorithm that has a totally different form from the given ones. }

\vspace{10pt}

\textcolor{Red}{1. First, describe your new algorithm and main steps in one sentence. The description must be inside within boxed \{\}.}

\vspace{10pt}

\textcolor{Red}{2. Next, implement the following Python function:
\textcolor{Blue}{\underline{\textbf{\$Program Template}}}} 

\vspace{10pt}

\textcolor{Red}{Do not give additional explanations.}

\vspace{10pt}

\end{adjustwidth}
\end{minipage}
}
\caption{An example of E1 prompt for TSP}
\label{fig: E1_prompt}
\end{figure}

\subsection{E2 Operator}

As shown in Figure~\ref{fig: E2_prompt}, the E2 operator is used to generate a new heuristic based on the common idea of the $5$ selected heuristics. 

\begin{figure}[H]
\centering
\Ovalbox{
\begin{minipage}{0.9\textwidth}
\vspace{10pt}

\begin{adjustwidth}{5pt}{}

\textcolor{Blue}{\underline{\textbf{\$Task Description}}} 

\vspace{10pt}

\textcolor{Red}{I have $5$ existing algorithms with their codes as follows: }

\textcolor{Red}{$<$Algorithm description$>$: } ...

\textcolor{Red}{$<$Code$>$: } ... 

\vspace{10pt}

...

\vspace{10pt}

\textcolor{Red}{Please help me create a new algorithm that has a totally different form from the given ones but can be motivated from them.}

\vspace{10pt}

\textcolor{Red}{1. Firstly, identify the common backbone idea in the provided algorithms.}

\vspace{10pt}

\textcolor{Red}{2. Secondly, based on the backbone idea describe your new algorithm in one sentence. The description must be inside within boxed \{\}.}

\vspace{10pt}

\textcolor{Red}{3. Thirdly, implement the following Python function:
\textcolor{Blue}{\underline{\textbf{\$Program Template}}}}

\vspace{10pt}

\textcolor{Red}{Do not give additional explanations.}

\vspace{10pt}

\end{adjustwidth}
\end{minipage}
}
\caption{An example of E2 prompt for TSP}
\label{fig: E2_prompt}
\end{figure}

\subsection{M1 Operator}

As shown in Figure~\ref{fig: M1_prompt}, the M1 operator is desired to generate a new heuristic based on a given heuristics to improve the performance. 

\begin{figure}[H]
\centering
\Ovalbox{
\begin{minipage}{0.9\textwidth}
\vspace{10pt}

\begin{adjustwidth}{5pt}{}

\textcolor{Blue}{\underline{\textbf{\$Task Description}}}

\vspace{10pt}

\textcolor{Red}{I have one algorithm with its code as follows:}

\textcolor{Red}{$<$Algorithm description$>$: } ...

\textcolor{Red}{$<$Code$>$: } ... 

\vspace{10pt}

\textcolor{Red}{Please assist me in creating a new algorithm that has a different form but can be a modified version of the algorithm provided.}

\vspace{10pt}

\textcolor{Red}{1. First, describe your new algorithm and main steps in one sentence. The description must be inside within boxed \{\}.}

\vspace{10pt}

\textcolor{Red}{2. Next, implement the following Python function:} \textcolor{Blue}{\underline{\textbf{\$Program Template}}}

\vspace{10pt}

\textcolor{Red}{Do not give additional explanations.}

\vspace{10pt}

\end{adjustwidth}
\end{minipage}
}
\caption{An example of M1 prompt for TSP}
\label{fig: M1_prompt}
\end{figure}

\subsection{M2 Operator}

As shown in Figure~\ref{fig: M2_prompt}, the goal of the M2 operator is to modify the parameters of a given heuristic. 

\begin{figure}[H]
\centering
\Ovalbox{
\begin{minipage}{0.9\textwidth}
\vspace{10pt}

\begin{adjustwidth}{5pt}{}

\textcolor{Blue}{\underline{\textbf{\$Task Description}}}

\vspace{10pt}

\textcolor{Red}{I have one algorithm with its code as follows:}

\textcolor{Red}{$<$Algorithm description$>$: } ...

\textcolor{Red}{$<$Code$>$: } ... 

\vspace{10pt}

\textcolor{Red}{Please identify the main algorithm parameters and assist me in creating a new algorithm that has a different parameter settings of the score function provided. }

\vspace{10pt}

\textcolor{Red}{1. First, describe your new algorithm and main steps in one sentence. The description must be inside within boxed \{\}. }

\vspace{10pt}

\textcolor{Red}{2. Next, implement the following Python function: }\textcolor{Blue}{\underline{\textbf{\$Program Template}}}

\vspace{10pt}

\textcolor{Red}{Do not give additional explanations.}

\vspace{10pt}

\end{adjustwidth}
\end{minipage}
}
\caption{An example of M2 prompt for TSP}
\label{fig: M2_prompt}
\end{figure}

\subsection{M3 Operator}

In Figure~\ref{fig: M3_prompt}, the M3 operator is used to simplify a given heuristic by eliminating redundant components. In this context, the task description and code requirements are not required. 

\begin{figure}[H]
\centering
\Ovalbox{
\begin{minipage}{0.9\textwidth}
\vspace{10pt}

\begin{adjustwidth}{5pt}{}

\textcolor{Red}{1. First, you need to identify the main components in the function below. }

\vspace{10pt}

\textcolor{Red}{2. Next, analyze whether any of these components can be overfit to the in-distribution instances. }

\vspace{10pt}

\textcolor{Red}{3. Then, based on your analysis, simplify the components to enhance the generalization to potential out-of-distribution instances. }

\vspace{10pt}

\textcolor{Red}{4. Finally, provide the revised code, keeping the function, inputs, and outputs unchanged. }

\vspace{10pt}

\textcolor{Red}{$<$Code$>$: } ... 

\vspace{10pt}

\textcolor{Red}{Do not give additional explanations.}

\vspace{10pt}

\end{adjustwidth}
\end{minipage}
}
\caption{An example of M3 prompt for TSP}
\label{fig: M3_prompt}
\end{figure}

\section{Designed Heuristics}~\label{sup:heu_example}

In this section, we present a variety of representative heuristics designed by LLM-based automated heuristic design frameworks, encompassing FunSearch~\cite{romera2024mathematical}, EoH~\cite{liu2024eoh}, and our own MEoH.

\subsection{BPP}

\subsubsection{EoH Heuristics}

The heuristic developed by EoH with the best performance in terms of the optimal gap, as shown in Figure~\ref{fig: best_EoH_BPP_heuristic}, utilizes sophisticated mathematical operators such as logarithm, square root, and exponential. The complexity of this scoring function renders it challenging to construct manually due to its intricate nature and reliance on advanced mathematical operations.

\begin{figure}[H]
\centering
\Ovalbox{
\begin{minipage}{0.9\textwidth}
\vspace{10pt}
\begin{adjustwidth}{5pt}{}
\textcolor{Green}{\textbf{Algorithm Description: }My new algorithm calculates the score for each bin as the sum of the bin's current capacity divided by the product of the logarithm of the difference between the bin's capacity and the item size and the square root of the difference between the bin's capacity and the item size, raised to the power of the bin's current capacity, and multiplied by the exponential function raised to the power of the item size multiplied by the difference between the bin's capacity and the item size. Additionally, the score is multiplied by the reciprocal of the bin's current capacity to prioritize bins with lower capacities.}

\vspace{10pt}

\begin{minted}[breaklines]{python}
import numpy as np

def score(item, bins):
    scores = (bins / ((np.log(bins - item) * np.sqrt(bins - item)) ** bins)) * np.exp(item * (bins - item)) * (1/bins)
    return scores
\end{minted}

\vspace{10pt}
\end{adjustwidth}
\end{minipage}
}
\caption{The EoH heuristic with the best optimal gap on BPP.}
\label{fig: best_EoH_BPP_heuristic}
\end{figure}

\subsubsection{FunSearch Heuristics}

The heuristic devised by FunSearch, illustrated in Figure~\ref{fig: best_FunSearch_BPP_heuristic}, it incorporates numerous sophisticated parameters and introduces a random noise. Unlike the EoH approach, the FunSearch heuristic relies on intricate parameter settings and stochastic perturbations for optimization.

\begin{figure}[H]
\centering
\Ovalbox{
\begin{minipage}{0.9\textwidth}
\vspace{10pt}
\begin{adjustwidth}{5pt}{}

\begin{minted}[breaklines]{python}
def priority(item: float, bins: np.ndarray) -> np.ndarray:
    eps = 1e-7

    # Calculate scores based on available space and current capacity
    scores = (bins - item) / (bins + eps)

    # Adjust the penalty if necessary
    penalty = np.power(np.min(bins), 0.5) * np.arange(len(bins)) * 0.01
    scores -= penalty

    # Scale the scores and add a weight
    weight = 0.8
    scores = (scores - np.min(scores)) / (np.max(scores) - np.min(scores)) * (1 - weight) + weight

    # Favor bins where the item fits perfectly
    scores += 0.5 * (bins == item)

    # Favor bins with relatively higher remaining capacity
    scores += 0.02 * (bins - item) / np.max(bins)

    # Normalize the priority values
    priority = scores / np.sum(scores)

    # Add a small randomness to the priorities for exploration
    priority += np.random.uniform(0, 1e-5, bins.shape) 

    # Handle the case where the sum of priorities is not equal to 1
    if np.abs(np.sum(priority) - 1) > 1e-6:
        remaining_capacity = bins - np.sum(priority * bins)
        priority += remaining_capacity / (np.sum(remaining_capacity) * len(bins))

    return priority

\end{minted}

\vspace{10pt}
\end{adjustwidth}
\end{minipage}
}
\caption{The FunSearch heuristic with the best optimal gap on BPP.}
\label{fig: best_FunSearch_BPP_heuristic}
\end{figure}
\newpage

\subsubsection{MEoH Heuristics}
\begin{figure*}
\centering
\includegraphics[width=\textwidth]{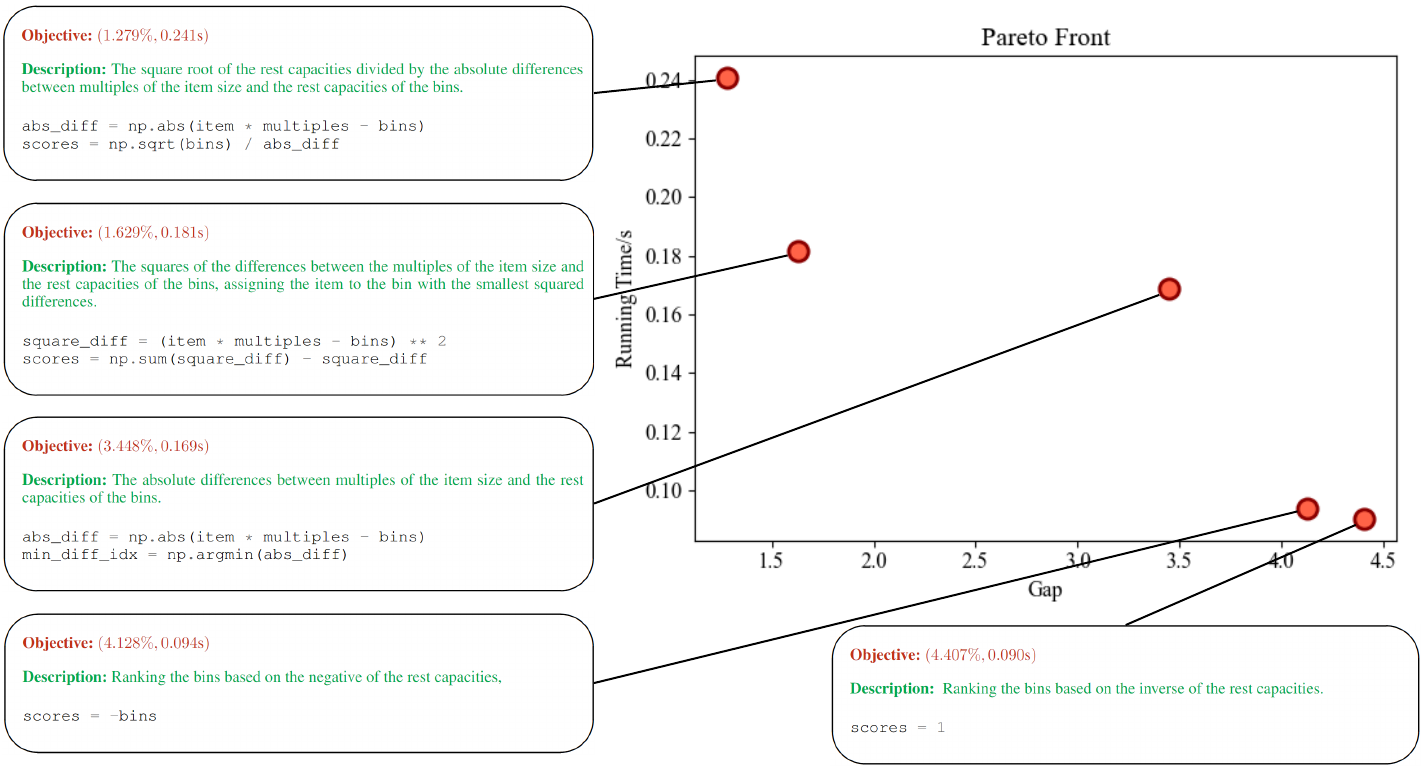}
\caption{An illustration of MEoH heuristics on BPP. }
\label{fig:BPP_PF_code}
\end{figure*}

In this section, Figure~\ref{fig:BPP_PF_code} showcases the heuristics developed by MEoH, featuring heuristic descriptions, corresponding code segments, and images in the objective space.

Specifically, these heuristics are designed to assign scores to bins based on the arriving items, subsequently arranging the items in bins with the highest scores.

Among these heuristics highlighted in Figure~\ref{fig:BPP_PF_code}, three exhibit superior performance in terms of the optimal gap, leveraging advanced mathematical operators like absolute value and square root. Furthermore, in the case of the fast heuristic, the score is consistently set to a fixed value of $1$, which deviates from the intended description.

Given the integration of these heuristics into a greedy algorithm, the running time demonstrates low variance. Nevertheless, these MEoH-generated heuristics effectively balance the optimal gap and running time, enabling adaptability to diverse scenarios.

\newpage

\subsection{TSP}

\subsubsection{EoH Heuristics}

The heuristic crafted by EoH, as illustrated in Figure~\ref{fig: best_EoH_TSP_heuristic}, intricately incorporates advanced mathematical functions such as tanh alongside sophisticated parameters. It is noteworthy that this complex operation is executed within two nested for-loops, resulting in a computational complexity of $\mathcal{O}(n^2)$. 

\begin{figure}[H]
\centering
\Ovalbox{
\begin{minipage}{0.9\textwidth}
\vspace{10pt}
\begin{adjustwidth}{5pt}{}
\textcolor{Green}{\textbf{Algorithm Description: }Update the edge distances in the edge distance matrix by applying a genetic algorithm-inspired method, where the update is determined by a combination of edge count, distance, usage, and a customized genetic function to promote global exploration and improved convergence.}

\vspace{10pt}

\begin{minted}[breaklines]{python}
import numpy as np

def update_edge_distance(edge_distance, local_opt_tour, edge_n_used):
    updated_edge_distance = np.copy(edge_distance)
    
    edge_count = np.zeros_like(edge_distance)
    for i in range(len(local_opt_tour) - 1):
        start = local_opt_tour[i]
        end = local_opt_tour[i + 1]
        edge_count[start][end] += 1
        edge_count[end][start] += 1
    
    edge_n_used_max = np.max(edge_n_used)
    mean_edge_distance = np.mean(edge_distance)
    
    for i in range(edge_distance.shape[0]):
        for j in range(edge_distance.shape[1]):
            if edge_count[i][j] > 0:
                score_factor = (np.tanh(edge_count[i][j]) / edge_count[i][j]) + (edge_distance[i][j] / mean_edge_distance) - (0.6 / edge_n_used_max) * edge_n_used[i][j]
                updated_edge_distance[i][j] += score_factor * (1 + edge_count[i][j])

    return updated_edge_distance
\end{minted}

\vspace{10pt}
\end{adjustwidth}
\end{minipage}
}
\caption{The EoH heuristic with the best optimal gap on TSP.}
\label{fig: best_EoH_TSP_heuristic}
\end{figure}

\subsubsection{FunSearch Heuristics}

The heuristic formulated by FunSearch, as depicted in Figure~\ref{fig: best_FunSearch_TSP_heuristic}, incorporates a logarithm operation base 2, Gaussian-distributed noise sampling, and intricate parameter configurations. It is worth noting that this heuristic only includes a single for-loop, indicating a computational efficiency that surpasses the aforementioned EoH heuristic. 

\begin{figure}[H]
\centering
\Ovalbox{
\begin{minipage}{0.9\textwidth}
\vspace{10pt}
\begin{adjustwidth}{5pt}{}

\begin{minted}[breaklines]{python}
def update_edge_distance(edge_distance: np.ndarray, local_opt_tour: np.ndarray, edge_n_used: np.ndarray) -> np.ndarray:
    num_nodes = edge_distance.shape[0]
    updated_edge_distance = np.copy(edge_distance)

    decay_factor = 0.99

    for i in range(num_nodes - 1):
        node_i, node_j = local_opt_tour[i], local_opt_tour[i + 1]

        edge_score = edge_distance[node_i, node_j] * np.log2((num_nodes - edge_n_used[node_i, node_j]) + 1)
        edge_score *= decay_factor ** edge_n_used[node_i, node_j]  # Multiply by decay factor
        edge_score += np.random.normal(0, 0.1)  # Add small noise

        updated_edge_distance[node_i, node_j] = edge_score
        updated_edge_distance[node_j, node_i] = edge_score

    return updated_edge_distance
\end{minted}

\vspace{10pt}
\end{adjustwidth}
\end{minipage}
}
\caption{The FunSearch heuristic with the best optimal gap on TSP.}
\label{fig: best_FunSearch_TSP_heuristic}
\end{figure}

\subsubsection{MEoH Heuristics}

\begin{figure*}
\centering
\includegraphics[width=\textwidth]{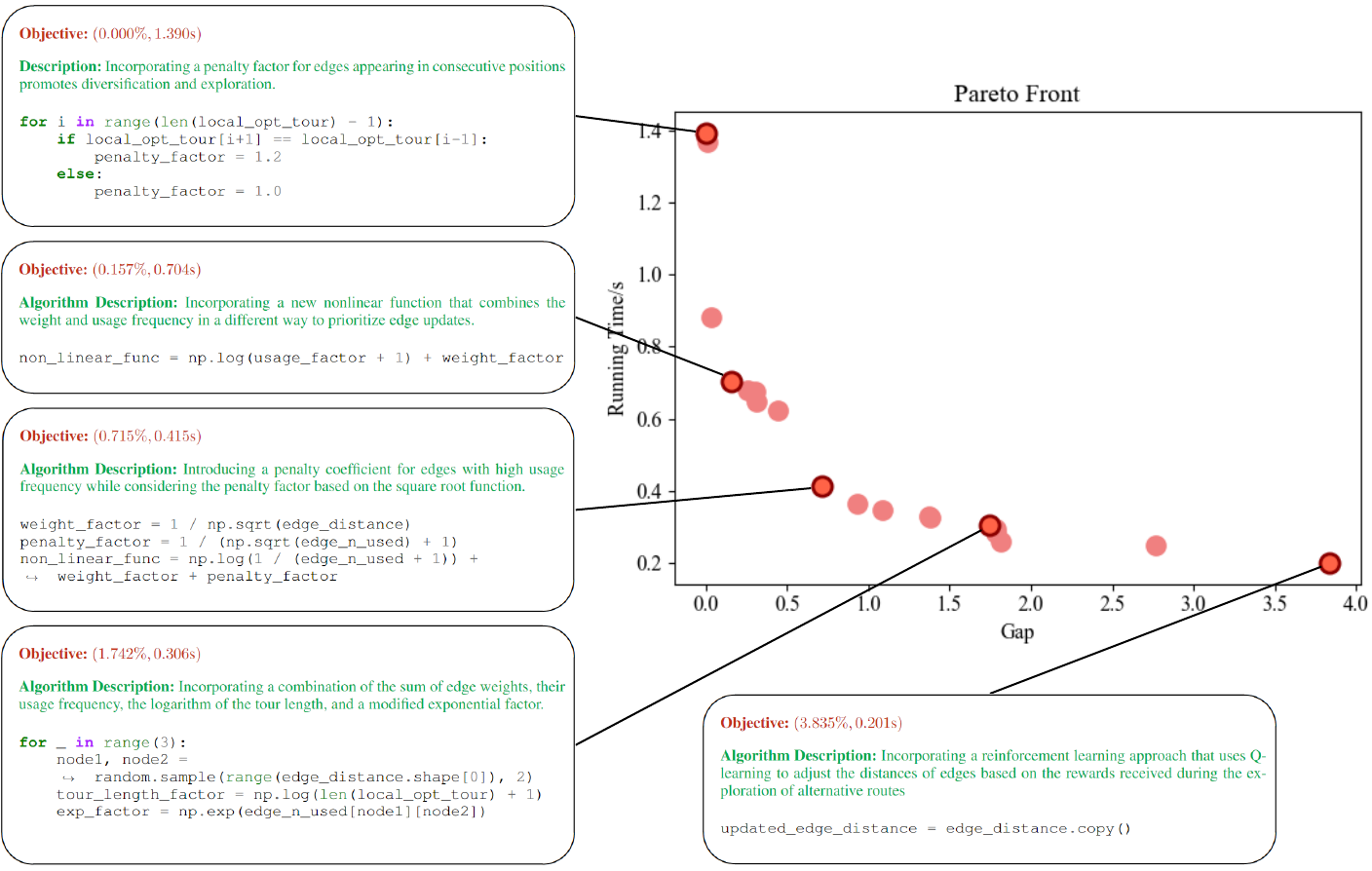}
\caption{An illustration of MEoH heuristics on TSP. }
\label{fig:TSP_PF_code}
\end{figure*}

In this section, the heuristics designed by MEoH are shown in Figure~\ref{fig:TSP_PF_code}, showcasing $5$ representative heuristic descriptions along with corresponding code segments, and visual representations of all the heuristics in the objective space.

In this work, GLS is employed to solve TSP, and the heuristics are designed to update the edge distance to facilitate the perturbation in each iteration.

As shown in Figure~\ref{fig:TSP_PF_code}, the designed heuristics leverage advanced mathematical operators including logarithm, square root, and exponential functions. Furthermore, for the fast heuristic, the edge distances remain unchanged, deviating from the original description due to the complexity of implementing Q-Learning.

These heuristics underscore the capability of our MEoH to strike a balance between the optimal gap and running time, allowing for effective adaptation to various scenarios.

\section{MEoH on 3-objective tasks}\label{sup:3o}

\begin{figure}[h!]
\centering
\subfloat[HV] {\includegraphics[width = 0.33\linewidth]{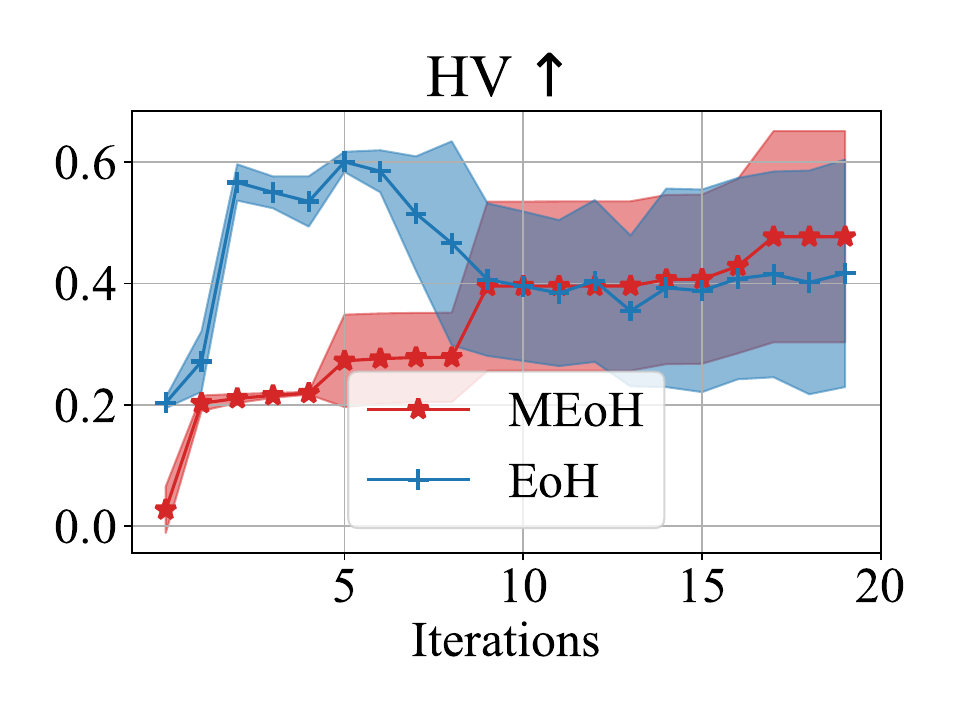}}
\subfloat[IGD]{\includegraphics[width = 0.33\linewidth]{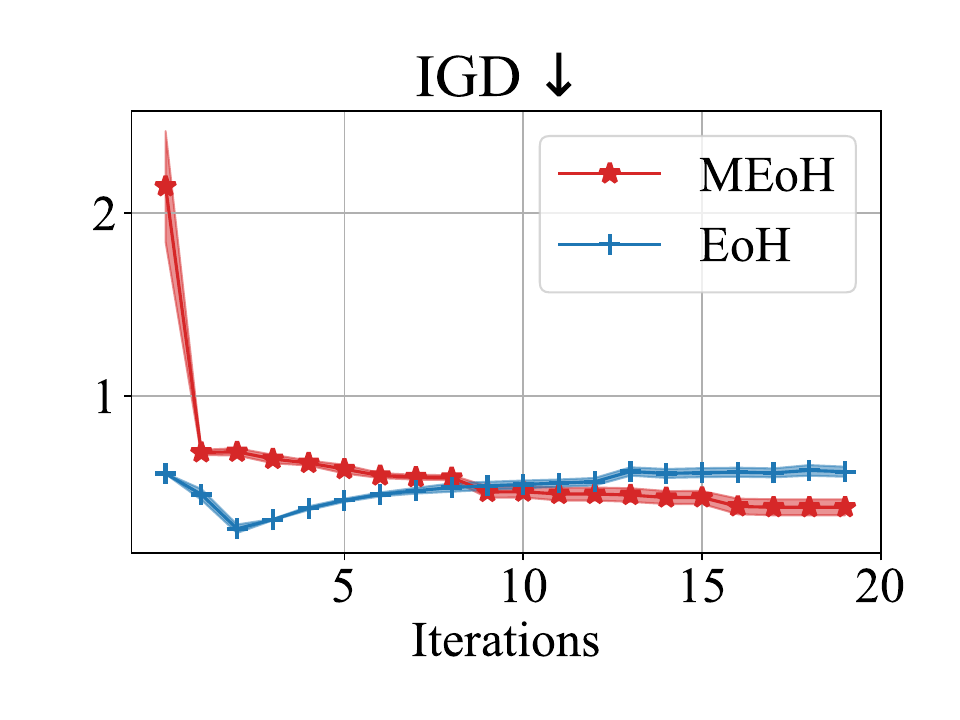}}
\caption{Comparations of EoH and MEoH on BPP5k.}
\label{fig: BPP_3o}
\end{figure}

\begin{figure}[h!]
\centering
\subfloat[HV] {\includegraphics[width = 0.33\linewidth]{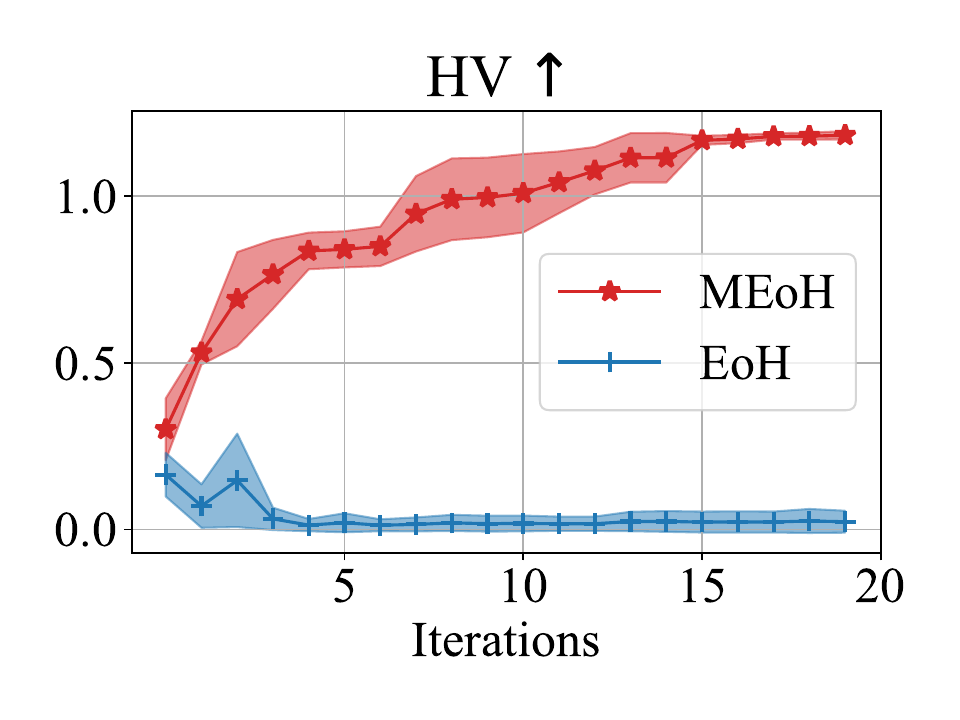}}
\subfloat[IGD]{\includegraphics[width = 0.33\linewidth]{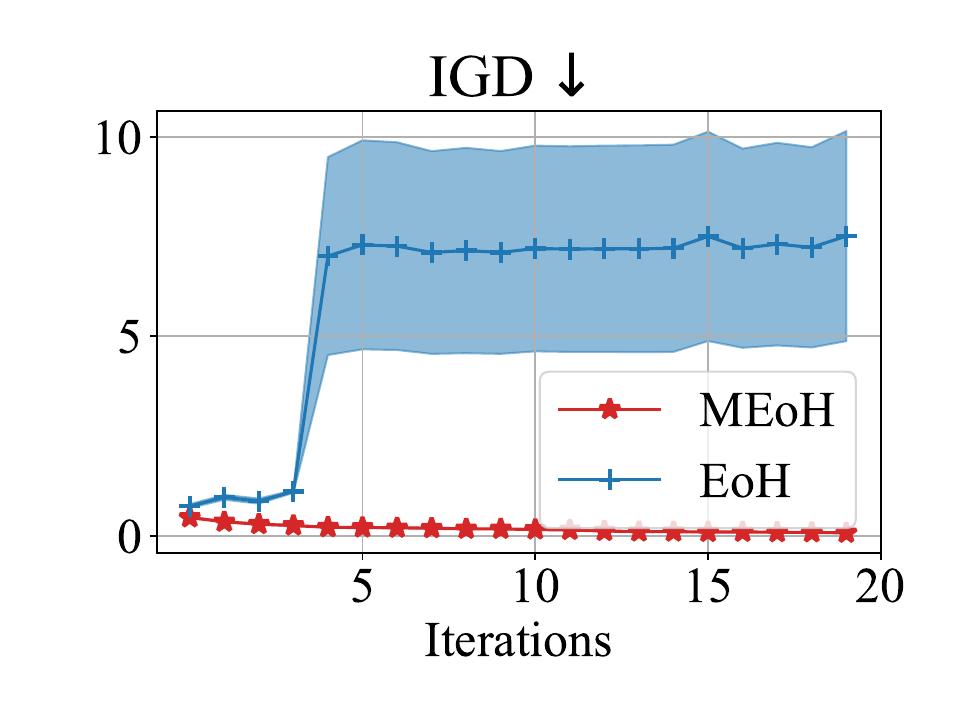}}
\caption{Comparations of EoH and MEoH on TSP100.}
\label{fig: TSP_GLS_3o}
\end{figure}

In this section, MEoH is utilized to develop heuristics while taking into account 3 objectives. In addition to performance and efficiency, we also consider code readability, which is crucial for user comprehension and maintenance of the programming code~(Buse et al., 2009). The readability is assessed using the Halstead difficulty metric~(Curtis et al., 1979). As illustrated in Figure~\ref{fig: BPP_3o} and~\ref{fig: TSP_GLS_3o}, MEoH continues to perform well, particularly in the TSP task. Despite the promising outcomes achieved with three objectives, future research will be essential to address the challenges associated with handling more objectives.

\end{document}